\documentclass[letterpaper]{article} 

\usepackage{aaai2027}

\usepackage[hyphens]{url}  
\usepackage{graphicx}      
\usepackage{natbib}        
\usepackage{caption}       
\usepackage[utf8]{inputenc}

\usepackage{amsmath}
\usepackage{amssymb}
\usepackage{amsfonts}
\usepackage{mathtools}
\usepackage{nicefrac}

\usepackage{booktabs}
\usepackage{multirow}
\usepackage{tabularx}
\usepackage{array}
\usepackage{makecell}
\usepackage{adjustbox}

\usepackage{subcaption}

\usepackage{microtype}
\usepackage{pifont}
\usepackage{enumitem}
\usepackage[table]{xcolor}

\usepackage{algorithm}
\usepackage{algorithmic}

\usepackage{newfloat}
\usepackage{listings}

\usepackage{hyperref}
\hypersetup{
    hidelinks
}

\newcommand{\ablationtablesetup}{%
    \fontsize{7.3pt}{8.6pt}\selectfont
    \setlength{\tabcolsep}{2.2pt}%
    \renewcommand{\arraystretch}{1.12}%
}

\newcolumntype{L}{>{\raggedright\arraybackslash}X}

\DeclareCaptionStyle{ruled}{
    labelfont=normalfont,
    labelsep=colon,
    strut=off
} 

\floatstyle{ruled}
\newfloat{listing}{tb}{lst}{}
\floatname{listing}{Listing}

\title{
WAM-Diff2: Hierarchical AR-to-Diffusion Distillation for Highly Efficient Autonomous Driving VLA
}

\author{
    Zhihao Zhu\textsuperscript{1}\thanks{Equal contribution},
    Hanlin Shang\textsuperscript{1}\footnotemark[1],
    Mingwang Xu\textsuperscript{1}\footnotemark[1],
    Feipeng Cai\textsuperscript{2}\footnotemark[1],
    Zhuolin He\textsuperscript{1},\\
    Yaoyi Li\textsuperscript{2},
    Jianhua Han\textsuperscript{2},
    Hang Xu\textsuperscript{2},
    Siyu Zhu\textsuperscript{1}\thanks{Corresponding author}
}

\affiliations{
    \textsuperscript{1}Fudan University,
    \textsuperscript{2}Yinwang Intelligent Technology Co., Ltd.\\
}

\makeatletter

\let\old@maketitle\@maketitle

\renewcommand{\@maketitle}{%
    \old@maketitle

    \vspace{-10.5mm}

    \noindent
    \makebox[\textwidth][c]{%
        {\small
        \href{https://ssssssuger.github.io/WAM-Diff2/}
        {\textbf{Project Page}}
        \quad
        $\vert$
        \quad
        \href{https://github.com/fudan-generative-vision/WAM-Diff2}
        {\textbf{Code}}
        }%
    }

    \par
    \vspace{2mm}

    \begin{center}

        \includegraphics[
            width=\textwidth
        ]{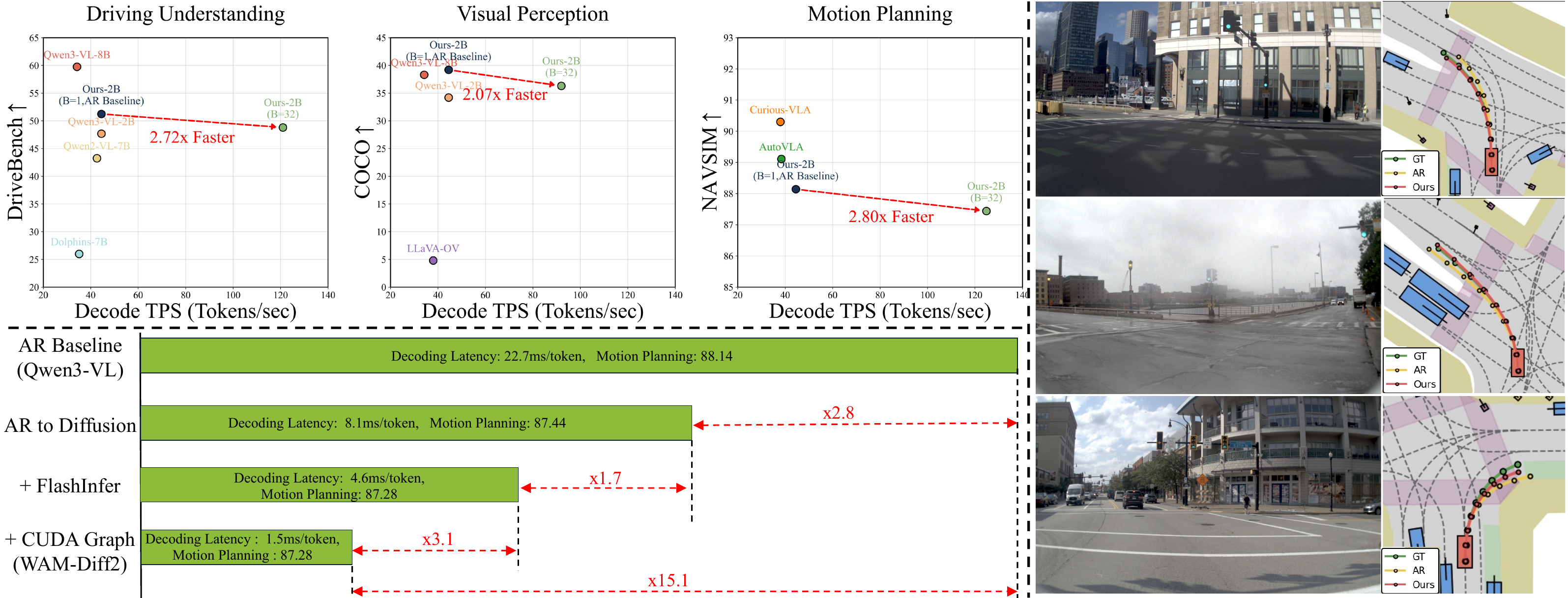}

        \captionof{figure}{
        Efficiency, multi-task performance, and planning robustness of WAM-Diff2.
        \textbf{Top left}:
        performance versus decoding throughput (tokens/s) across driving understanding
        (DriveBench), visual perception (COCO), and motion planning (NAVSIM).
        Our parallel discrete diffusion approach ($B=32$) yields up to a
        2.8$\times$ speedup over the autoregressive baseline while preserving
        high multi-task proficiency.
        \textbf{Bottom left}:
        latency optimization breakdown.
        Transitioning from the AR baseline to the WAM-Diff2 diffusion paradigm
        accelerates decoding by 2.8$\times$.
        Integrating FlashInfer and CUDA Graphs yields a cumulative
        15.1$\times$ latency reduction
        (from 22.7~ms/token to 1.5~ms/token)
        with negligible planning degradation.
        \textbf{Right}:
        mitigation of exposure bias in long-horizon trajectory prediction.
        Compared to the AR baseline, which suffers from compounding drift errors,
        WAM-Diff2 delivers highly robust, temporally consistent planning that
        closely aligns with the ground truth.
        }

        \label{fig:teaser}

    \end{center}
}

\makeatother

\begin{document}

\maketitle

\begin{abstract}
Vision-Language-Action (VLA) models have emerged as a prominent paradigm for end-to-end autonomous driving; 
however, their efficient deployment is severely constrained by high computational latency and exposure bias arising from sequential autoregressive decoding. 
Conversely, while specialized diffusion policies enable low-latency, 
parallel execution, 
training them from scratch typically yields narrow, 
single-task architectures that lack holistic visual-linguistic reasoning. 
Successfully transforming pre-trained autoregressive generalists into parallel diffusion models could combine multi-task cognitive intelligence with execution efficiency, 
yet this transition presents a formidable architectural challenge due to mismatched attention patterns (causal versus bidirectional) and divergent optimization objectives. 
To bridge this divide, 
we introduce WAM-Diff2, 
a multi-task discrete diffusion VLA framework powered by a three-stage hierarchical distillation strategy. 
By structuring the architectural shift through progressive block-wise adaptation, 
block-wise distillation, 
and model-wise cross-scale distillation, 
WAM-Diff2 preserves the underlying semantic foundations of the base model while accelerating inference. 
Extensive evaluations across driving understanding, perception, and planning benchmarks demonstrate that WAM-Diff2 effectively mitigates exposure bias and achieves performance parity with autoregressive baselines. 
Crucially, the autoregressive-to-diffusion transition yields a $2.8\times$ decoding speedup, 
which scales to an ultimate $15.1\times$ acceleration when combined with system-level optimizations including FlashInfer and CUDA Graphs. Our project code will soon be open-sourced.
\end{abstract}

\begin{figure*}[!t]
    \centering
    \includegraphics[width=1.0 \linewidth]{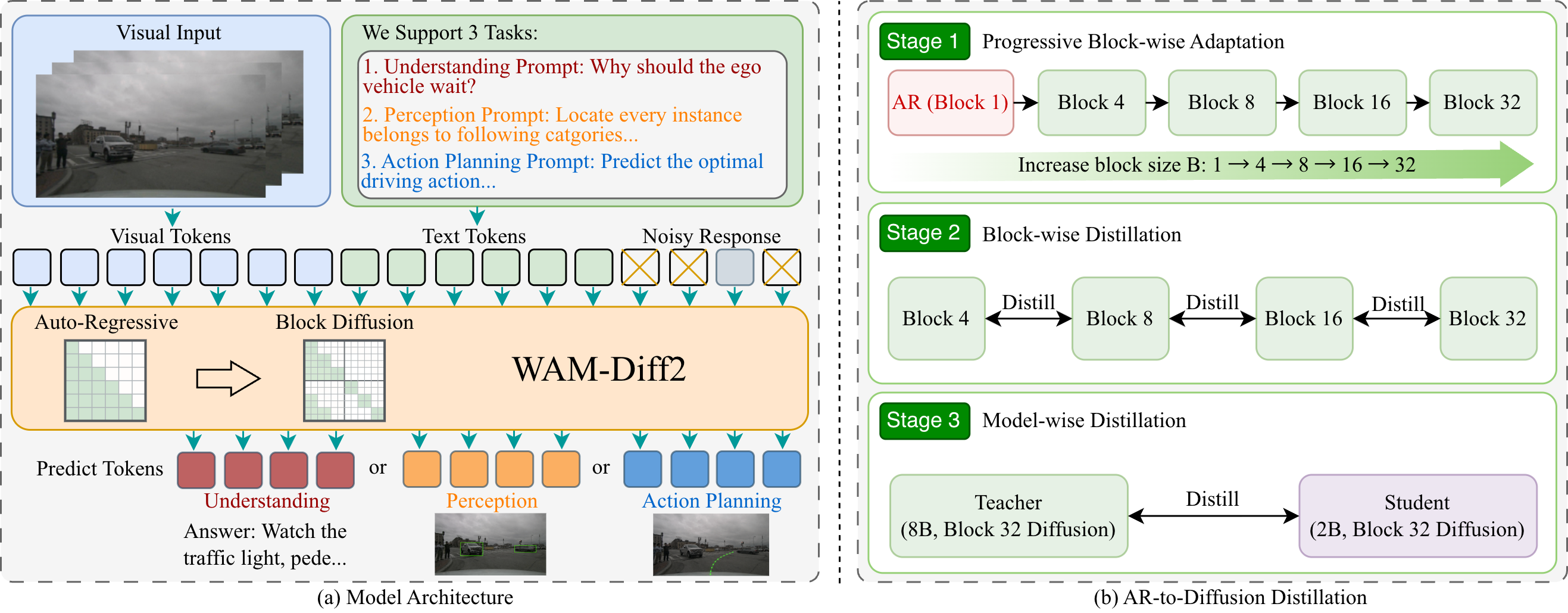}
    \vspace{-7mm}
    \caption{Architectural overview of WAM-Diff2 and the three-stage hierarchical distillation framework. 
    (a) Multi-view images, linguistic instructions, and noisy response tokens are mapped into a unified token space for multi-task inference across driving understanding, perception, and action planning. 
    By replacing traditional autoregressive causal masks with a block-causal attention pattern, 
    WAM-Diff2 enables bidirectional parallel token refinement within each decoding block while preserving cross-block causal constraints. 
    (b) The autoregressive-to-diffusion transition is accomplished through three progressive stages: 
    Stage 1 incrementally expands the decoding block size ($B = 1 \rightarrow 32$) to smoothly adapt the underlying attention mechanisms; 
    Stage 2 applies recursive block-wise distillation from a stable, small-block diffusion teacher to guide larger-block student variants; 
    and Stage 3 employs model-wise cross-scale distillation to transfer holistic semantic capabilities from an advanced 8B diffusion teacher to a highly efficient 2B student.}
\vspace{-4mm}
\label{fig:overview}
\end{figure*}

\section{Introduction}
\label{sec:intro}

Vision-Language-Action models~\cite{sima2024drivelm,hwang2024emma,li2025drivevlaw0,chen2025drivinggpt,li2025recogdrive,han2025percept,li2025reflectdrive,cui2025vilad,zhou2025opendrivevla,zhou2025autovla,li2026unidrivevla} have emerged as the leading paradigm for unified end-to-end autonomous driving, 
consolidating diverse tasks--ranging from linguistic scene understanding~\cite{li2024llava,bai2025qwen3,han2025percept,li2025recogdrive} to 2D/3D perception~\cite{han2025percept,li2026unidrivevla,bai2025qwen3} and motion planning~\cite{wu2022trajectory,hu2023planning,jiang2023vad,transfuser,li2025recogdrive,wang2025diffad,li2025drivevlaw0,diffusiondrivev2,kim2026safedrive}--within a single network architecture. 
Despite their impressive multi-task proficiency, prevailing driving generalists are predominantly built upon causal autoregressive foundations. This design choice imposes fundamental limitations on their operational deployment. Structurally, the sequential nature of autoregressive decoding generates outputs token-by-token, incurring high computational latency that threatens real-time driving safety. Algorithmically, autoregressive training relies on ground-truth prefixes via teacher forcing~\cite{cen2024bridging,gu2023minillm,agarwal2024policy,bachmann2024pitfalls}, leaving the model inherently vulnerable to compounding errors during long-horizon inference—a phenomenon widely recognized as exposure bias.

To overcome these latency and robustness bottlenecks, 
the community has bifurcated into two disconnected lines of research. 
On one side, autoregressive generalists~\cite{hwang2024emma,li2025recogdrive,chen2025drivinggpt,han2025percept,zhou2025autovla} leverage large language model scales to achieve broad multi-task semantic reasoning at the expense of computational efficiency and causal fragility. 
On the opposite side, 
planning-centric diffusion models~\cite{zheng2025diffusion,liao2025diffusiondrive,diffusiondrivev2,xu2025wam} utilize highly parallel, 
discrete or continuous diffusion processes to achieve low-latency control and robust bidirectional trajectory refinement. 
However, training these specialized architectures from scratch typically yields narrow, 
single-task systems that lack the holistic understanding and visual-linguistic reasoning capabilities of their autoregressive counterparts. 
Consequently, existing paradigms present a rigid compromise between multi-task cognitive intelligence and parallelized execution efficiency.

In this work, 
we transcend this dichotomy by introducing a universal architecture translation paradigm rather than presenting a localized, planning-only diffusion model. 
We argue that the vast semantic knowledge and multi-task reasoning embedded within mature, 
pre-trained autoregressive models~\cite{li2024llava,bai2025qwen3,zhu2025internvl3exploringadvancedtraining} should be leveraged, 
rather than discarded, 
to initialize parallel generation architectures. 
The core challenge lies in the radical paradigm gap between sequential, 
unidirectional causal pre-training and parallel, 
bidirectional context refinement. 
This transition presents a formidable architectural challenge due to mismatched attention patterns and divergent optimization objectives.

To bridge this fundamental divide, 
we present WAM-Diff2, 
a multi-task discrete diffusion VLA framework powered by a novel three-stage hierarchical distillation strategy. 
Instead of initiating training from scratch, our framework seamlessly transforms any advanced, pre-trained autoregressive driving generalist into an ultra-efficient parallel refinement architecture. 
Our core insight is that the architectural transition must be staged to protect the underlying semantic foundations. 
In Stage I (Progressive Block-Wise Adaptation), 
we design a block-causal attention mechanism that incrementally relaxes rigid causal constraints through a curriculum-based expansion of decoding blocks, 
ensuring mathematical stability during the attention shift. 
In Stage II (Block-Wise Distillation), 
we actively expose the student to intermediate noisy states generated during parallel decoding, 
utilizing an on-policy~\cite{agarwal2024policy,gu2023minillm}, 
small-block diffusion teacher to enforce paradigm consistency and eliminate exposure bias. 
In Stage III (Model-Wise Cross-Scale Distillation), 
we address the capacity-efficiency trade-off by scaling knowledge transfer from an advanced 8B diffusion teacher to a compact 2B student. 
Crucially, we discover that models sharing the same diffusion paradigm exhibit highly aligned token prediction patterns, 
allowing the smaller model to easily inherit complex reasoning from the larger teacher without compromising parallel decoding speed.

Comprehensive evaluations across trajectory planning (NAVSIM~\cite{dauner2024navsim}, Bench2Drive~\cite{jia2024bench2drive}), 
driving-oriented visual question answering (DriveBench~\cite{xie2025vlms}, LingoQA~\cite{marcu2024lingoqa}), 
and visual perception (COCO~\cite{lin2014microsoft}) benchmarks demonstrate that WAM-Diff2 achieves strict performance parity or superiority relative to its autoregressive foundation while delivering an obvious inference speedup. 
Specifically, the autoregressive-to-diffusion transition yields a $2.8\times$ decoding speedup, 
which scales to an ultimate $15.1\times$ acceleration when combined with system-level optimizations including FlashInfer and CUDA Graphs. 
Ultimately, 
our framework establishes a highly scalable, 
low-cost pipeline capable of transforming existing and future state-of-the-art autoregressive VLAs into high-throughput, highly efficient deployable diffusion agents.

\section{Related Works}
\paragraph{VLA for Autonomous Driving.}
Vision-Language-Action models have evolved from foundational scene understanding and VQA~\cite{xu2024drivegpt4, tian2024drivevlm, sima2024drivelm, ma2024dolphins, nie2024reason2drive, wang2025omnidrive} toward unified end-to-end control~\cite{zhou2025opendrivevla, zhou2025autovla, fu2025orion, mao2023gpt, shao2024lmdrive, jiang2024senna, xu2025drivegpt4}. 
Current driving generalists, 
such as EMMA~\cite{hwang2024emma}, Percept-WAM~\cite{han2025percept}, DriveMoE~\cite{yang2025drivemoe}, and UniDriveVLA~\cite{li2026unidrivevla}, 
integrate perception and planning within autoregressive frameworks~\cite{mao2023gpt, chen2025drivinggpt, zhou2025opendrivevla, zhou2025autovla, li2026unidrivevla}. 
However, these sequential architectures are inherently limited by unidirectional causal constraints, high inference latency, and exposure bias.

\paragraph{Discrete Diffusion Models.}
Discrete diffusion provides a robust alternative for sequential generation~\cite{austin2021structured, hoogeboom2021argmax, campbell2022continuous, li2022diffusion, gong2022diffuseq, lou2023discrete, shi2024simplified, sahoo2024simple, gat2024discrete}. 
Recently, masked diffusion has scaled to large-scale language and multimodal domains~\cite{nie2025large, ye2025dream, you2025llada, yang2025mmada, ou2024your, arriola2025block, von2025generalized}, 
enabling parallel iterative refinement. 
While applied to trajectory generation in robotics and driving~\cite{cui2025vilad, li2025reflectdrive, xu2025wam, dang2026drivefine, zhang2026efficient, liang2025discrete, wen2025llada, ye2025dream, wen2025dvla, liu2026mmada, chen2026dfm}, 
existing diffusion models often lack the multi-task breadth of autoregressive counterparts. We bridge this capability gap by transforming a unified autoregressive VLA into a discrete diffusion framework.

\paragraph{LLM and LVM Distillation.} 
Knowledge distillation~\cite{hinton2015distilling} is critical for model compression. 
Traditional off-policy distillation for autoregressive models~\cite{kim2016sequence, sanh2019distilbert} suffers from exposure bias~\cite{bengio2015scheduled, ranzato2015sequence}, 
which recent on-policy approaches~\cite{agarwal2024policy, li2026rethinking, afsharrad2026policy} and divergence-based objectives~\cite{gu2023minillm} seek to mitigate. 
Transferring knowledge from autoregressive teachers to diffusion students is uniquely challenging due to the discrepancy between causal and bidirectional attention. 
Our tiered framework addresses this through progressive adaptation and cross-scale alignment.
\section{Methodology}

\paragraph{Overview.} 
We propose a VLA architecture that seamlessly transitions from standard autoregressive sequence generation to highly parallel discrete diffusion. 
This section first formalizes the problem by contrasting the autoregressive foundation with our discrete diffusion formulation (Section~\ref{sec:preliminary}). 
We then introduce our core contribution: 
a three-stage distillation framework designed to transfer the multi-task autoregressive VLA teacher to a highly efficient, 
parallel diffusion student (Section~\ref{sec:distillation}). 
Finally, we detail the network architecture and training protocols (Section~\ref{sec:network}).

\subsection{Preliminaries}
\label{sec:preliminary}
\paragraph{VLA Formulation.}
Our baseline model 
$\Phi$ is a multi-task VLA generalist that maps multimodal sensory streams and linguistic instructions into a unified discrete output space. 
Given a set of $N_c$ surrounding multi-camera RGB images $\mathcal{I}^{\text{cam}} = \{I_k\}_{k=1}^{N_c}$ and a natural-language task instruction $\mathcal{L}^{\text{instruct}}$, the network generates a task-conditioned sequence:
$\mathcal{O} = \Phi(\mathcal{I}^{\text{cam}}, \mathcal{L}^{\text{instruct}})$.
We represent all multi-task targets using a shared, unified text tokenizer, 
deliberately omitting task-specific projection heads.
This framework encapsulates three core autonomous driving capabilities:
\textit{(1) Driving Understanding}: 
Given a linguistic query $\mathcal{L}^{\text{qa}}$, 
the agent aligns visual features with text semantics to produce a natural-language scene-reasoning response $\mathcal{O}^{\text{qa}}$.
\textit{(2) Visual Perception}:
For spatial grounding tasks, the model outputs a serialized sequence of $M$ bounding boxes $\mathcal{O}^{\text{det}} = \{(b_i, c_i, s_i)\}_{i=1}^{M}$, 
where $b_i \in \mathbb{R}^4$ denotes bounding box coordinates, 
$c_i$ is the class label, and $s_i$ represents the confidence score.
\textit{(3) Motion Planning}:
Conditioned on a navigation command $\mathcal{L}^{\text{nav}}$ and the ego-vehicle status $\mathcal{L}^{\text{ego}}$, 
the network generates a future trajectory $\mathcal{O}^{\text{traj}} = \{(\hat{x}_t, \hat{y}_t)\}_{t=1}^{T}$ comprising $T$ BEV waypoints.

\paragraph{Autoregressive VLA.}
Conventional driving generalists serialize these multi-task outputs into a flat sequence of discrete tokens 
$\mathcal{O} = (o_1, \dots, o_L)$. 
The network optimizes the parameters by maximizing the joint likelihood via a standard causal language modeling objective:
\begin{equation}
p(\mathcal{O} \mid \mathcal{I}^{\text{cam}}, \mathcal{L}^{\text{instruct}}) = \prod_{i=1}^{L} p(o_i \mid o_{<i}, \mathcal{I}^{\text{cam}}, \mathcal{L}^{\text{instruct}})
\end{equation}
where $o_{<i}$ represents the causal prefix history. 
While this causal formulation leverages the scaling laws of pre-trained vision-language models, 
its sequential token-by-token generation incurs a computational complexity of $\mathcal{O}(L)$. 
This linear dependency presents two fundamental limitations: 
it introduces severe latency bottlenecks that impede real-time deployment, 
and its reliance on ground-truth prefixes during training (teacher forcing~\cite{song2020lightpaff}) induces exposure bias, 
leaving the system susceptible to compounding drift errors during closed-loop execution.

\paragraph{Discrete Diffusion VLA.} 
To bypass these computational and architectural bottlenecks, 
we reformulate sequence generation as an iterative denoising process over a finite discrete state space $\mathcal{V}$. 
Given a clean target sequence $\mathbf{o}^0 = (o_1, \dots, o_L) \in \mathcal{V}^L$, 
we employ an absorbing-state (masked) discrete diffusion formulation defined by a forward and reverse process.

\textit{Forward Process:}
The forward process progressively corrupts the clean token sequence $\mathbf{o}^0$ into a noisy state $\mathbf{o}^t$ over $T$ discrete timesteps. 
At each transition step, a token either retains its identity or transitions to a special absorbing $\text{[MASK]}$ token. 
Representing tokens as one-hot vectors, the transition matrix $Q_t$ is defined as:
\begin{equation}
q(\mathbf{o}^t \mid \mathbf{o}^{t-1}) = \prod_{i=1}^{L} \mathbf{o}_i^{t-1} Q_t,\; Q_t = (1 - \beta_t) \mathbf{I} + \beta_t \mathbf{1} e_m^\top,
\end{equation}
where $e_m$ is the one-hot base vector for the $[\text{MASK}]$ token, 
$\mathbf{I}$ is the identity matrix, 
and $\beta_t \in [0,1]$ defines the noise schedule. 
This formulation allows the intermediate marginal distribution to be evaluated in closed form as $q(\mathbf{o}^t \mid \mathbf{o}^0) = \mathbf{o}^0 \bar{Q}_t$, 
where $\bar{Q}_t = \prod_{s=1}^{t} Q_s$, 
facilitating highly efficient parallel training via random state sampling.

\textit{Reverse Parallel Refinement:} 
The denoising network $\Phi_\theta$ is trained to reconstruct the clean sequence directly from the corrupted observation, 
modeling the conditional distribution $p_\theta(\mathbf{o}^0 \mid \mathbf{o}^t, c)$. 
During inference, 
we execute a parallel remasking strategy: 
starting from a fully masked sequence, the model predicts all token distributions simultaneously. 
At each denoising step, it fixes a subset of tokens with the highest prediction confidence and re-masks the remaining tokens, compressing full-sequence decoding complexity to $\mathcal{O}(T)$ steps, where $T \ll L$.

\begin{table*}[!t]
\centering
\caption{
Multi-task performance and decoding efficiency across driving generalist paradigms.
We evaluate a single unified model checkpoint across three core autonomous driving capabilities: driving understanding (DriveBench, LingoQA), 
visual perception (COCO mAP), 
and motion planning (NAVSIM v1/v2 PDMS/EPDMS). 
Efficiency is reported as decoding throughput (Tokens Per Second, TPS) under identical hardware configurations, 
where the dual values for our diffusion models denote standard implementation versus hardware-optimized execution (FlashInfer + CUDA Graphs). 
The $^\dagger$ and $^*$ symbols denote external methods utilizing reinforcement learning and score-based candidate selection, respectively. 
}
\label{tab:main_results}
\setlength{\tabcolsep}{4pt}
\renewcommand{\arraystretch}{1.12}
\footnotesize

\resizebox{\textwidth}{!}{
\begin{tabular}{lcccccc}
\toprule
\multirow{2}{*}{\textbf{Method}}
& \multicolumn{2}{c}{\textbf{Driving Understanding}}
& \multicolumn{1}{c}{\textbf{Visual Perception}}
& \multicolumn{2}{c}{\textbf{Motion Planning}}
& \multicolumn{1}{c}{\textbf{Efficiency}} \\
\cmidrule(lr){2-3}
\cmidrule(lr){4-4}
\cmidrule(lr){5-6}
\cmidrule(lr){7-7}
& \textbf{DriveBench} $\uparrow$
& \textbf{LingoQA} $\uparrow$
& \textbf{COCO} $\uparrow$
& \textbf{NAVSIM V1} $\uparrow$
& \textbf{NAVSIM V2} $\uparrow$
& \textbf{Decode TPS} $\uparrow$ \\
\midrule

\multicolumn{7}{c}{\textbf{Multi-Task VLMs and VLAs}} \\
\midrule

Qwen3-VL-2B~\cite{bai2025qwen3}
& 47.68 & 48.00 & 34.20 & -- & -- & 44.5 \\

Qwen3-VL-8B~\cite{bai2025qwen3}
& \textbf{59.75} & 60.20 & 38.30 & -- & -- & 34.2 \\

Qwen2-VL-7B~\cite{wang2024qwen2}
& 43.25 & 52.60 & 19.30 & -- & -- & 42.6 \\

Qwen2.5-VL-3B~\cite{bai2025qwen25vltechnicalreport}
& 30.52 & 52.40 & 18.70 & -- & -- & 37.6 \\

Qwen2.5-VL-7B~\cite{bai2025qwen25vltechnicalreport}
& 35.55 & 60.00 & 20.10 & -- & -- & 40.2 \\

InternVL3-2B~\cite{zhu2025internvl3exploringadvancedtraining}
& 33.30 & 49.40 & 7.50 & -- & -- & 45.9 \\

InternVL3-8B~\cite{zhu2025internvl3exploringadvancedtraining}
& 37.99 & 45.00 & 17.50 & -- & -- & 42.5 \\

LLaVA-OV~\cite{li2024llava}
& 21.99 & 38.80 & 4.80 & -- & -- & 38.0 \\

Percept-WAM~\cite{han2025percept}
& -- & -- & \textbf{51.70}
& 80.40 / 90.20$^*$
& -- & -- \\

Recogdrive-8B~\cite{li2025recogdrive}
& \underline{56.71} & 67.20 & --
& 86.50 / 89.60$^\dagger$
& -- & -- \\

\midrule

\textbf{Ours-2B ($B=1$, AR Baseline)}
& 51.23
& \textbf{68.40}
& \underline{39.20}
& \textbf{88.14} / \textbf{91.50}$^*$
& \textbf{88.32}
& 44.5 \\

\textbf{Ours-2B ($B=4$)}
& 48.92
& \underline{68.20}
& 37.80
& \underline{87.87} / \underline{91.47}$^*$
& \underline{87.99}
& 68.3 / 401.4 \\

\textbf{Ours-2B ($B=8$)}
& 48.81
& 68.00
& 37.30
& 87.57 / 91.19$^*$
& 87.78
& 89.5 / 561.6 \\

\textbf{Ours-2B ($B=16$)}
& 48.77
& 66.00
& 36.50
& 87.32 / 91.26$^*$
& 87.30
& \underline{108.4} / \underline{635.3} \\

\textbf{Ours-2B ($B=32$)}
& 48.80
& 65.80
& 36.30
& 87.44 / 91.05$^*$
& 87.50
& \textbf{124.8} / \textbf{673.4} \\

\bottomrule
\end{tabular}
}
\end{table*}

\subsection{AR-to-Diffusion VLA Transformation}
\label{sec:distillation}

Transferring semantic knowledge from a causal autoregressive teacher to a bidirectional diffusion student introduces a fundamental paradigm gap,
which arises from mismatched attention patterns (causal vs. bidirectional) and divergent optimization objectives (next-token prediction vs. iterative denoising). 
To bridge this gap without performance degradation, we propose a three-stage hierarchical distillation framework.

\paragraph{Stage I: Progressive Block-wise Adaptation.}
We formulate the architectural transition from sequential decoding to parallel generation as a curriculum learning problem. 
We define the autoregressive baseline as a limiting case of block-wise generation where the decoding block size $B = 1$. 
To smoothly transition to global bidirectional attention, 
we define a curriculum that incrementally expands the decoding block size $B \in \{4, 8, 16, 32\}$.

For a target block size $B$, 
the sequence $\mathbf{y}$ is partitioned into non-overlapping blocks. 
The student network $p_{\theta_B}$ is optimized to reconstruct a fully masked block $\mathcal{B} \subset \mathbf{y}$ conditioned on the visual context $x$, 
the unmasked context $\mathbf{y}_{\setminus \mathcal{B}}$, and the discrete diffusion timestep $t$:
\begin{equation}
\mathcal{L}_{\mathrm{SFT}}^{B} = -\mathbb{E}_{t, \mathcal{B}:\vert{}\mathcal{B}\vert{}=B} \sum_{i \in \mathcal{B}} \log p_{\theta_B} (y_i \mid x, \mathbf{y}_{\setminus \mathcal{B}}, t).
\end{equation}
To prevent semantic degradation during this attention shift, we employ a progressive bootstrapping initialization scheme:
\begin{equation}
\theta_{B}^{(0)} \leftarrow \theta_{B/2}^{\mathrm{SFT}}, \quad \text{where } \theta_{1} = \theta_{\mathrm{AR}}.
\end{equation}
By sequentially expanding $B$, 
the network progressively internalizes joint token distributions and stabilizes the transition from causal to block-causal attention mechanisms.

\paragraph{Stage II: Block-wise Distillation.}
Although Stage I structurally adapts the attention layers to larger block dimensions, 
optimizing exclusively via ground-truth targets provides poor guidance for iterative refinement trajectory. 
To instill robust parallel denoising mechanics, 
Stage II introduces block-wise distillation from a stable, small-block diffusion teacher.
We employ progressive block-wise distillation following the sequence $4 \rightarrow 8 \rightarrow 16 \rightarrow 32$.
Specifically, the block-4 model $p_{\theta_4}$ obtained in Stage I is used to distill the block-8 model $p_{\theta_8}$.
The resulting block-8 model subsequently serves as the teacher for the block-16 model, and the same procedure is repeated for the block-32 model.
At each stage, the student $p_{\theta_B}$ is trained to match the predictive distribution of the preceding teacher $p_{\theta_{B/2}}$ over identical corrupted states.
To prevent the mode-covering behavior associated with standard forward KL divergence--which often compromises multi-modal trajectory distributions--we align the output distributions across the target block ${B}$ using a symmetric Jensen-Shannon Divergence (JSD) loss~\cite{englesson2021generalized}:
\begin{equation}
\resizebox{1.0\columnwidth}{!}{$
\displaystyle
\mathcal{L}_{\mathrm{BWD}}^{B} = \mathbb{E}_{t, \mathcal{B}, \tilde{\mathbf{y}}^{B}_{t}} \sum_{i \in \mathcal{B}} \mathrm{JSD} \left( p_{\theta_{B/2}}^{(i)} (\cdot \mid x, \tilde{\mathbf{y}}^{B}_{t}, t) \parallel p_{\theta_B}^{(i)} (\cdot \mid x, \tilde{\mathbf{y}}^{B}_{t}, t) \right),
$}
\end{equation}
where $\tilde{\mathbf{y}}^{B}_{t}$ represents the noisy token sequence, 
and the JSD between two distributions $p$ and $q$ is defined as:
\begin{equation}
\mathrm{JSD}(p \parallel q) = \frac{1}{2} \mathrm{KL}\left(p \parallel \frac{p+q}{2}\right) + \frac{1}{2} \mathrm{KL}\left(q \parallel \frac{p+q}{2}\right).
\end{equation}
Distilling from an intermediate diffusion anchor rather than the raw autoregressive baseline encourages that the optimization signal remains strictly paradigm-consistent, 
isolating the learning objective to parallel token scaling rather than cross-paradigm reconciliation.

\paragraph{Stage III: Model-Wise Cross-Scale Distillation.}
To recover the multi-task high-level intelligence lost during the compression to a 2B student, 
we scale the model capacity using a large-scale (8B) teacher. 
We observe that knowledge transfer is most effective when the teacher and student share consistent ``thinking patterns'', 
which we quantify using the top-$K$ overlap ratio $\rho_K$:
\begin{equation}
\rho_K(p_T, p_S) = \frac{|\text{TopK}(p_T) \cap \text{TopK}(p_S)|}{K}
\end{equation}
Our empirical analysis shows that $\rho_K$ is significantly higher between two diffusion models than between an autoregressive teacher and a diffusion student. 
Therefore, we utilize an 8B block-32 diffusion model $p_{\phi_{32}}^{8B}$ as our ultimate teacher. 
The 2B student $p_{\theta_B}^{2B}$ ($B=32$) is trained to match this 8B teacher on its own trajectories:
\begin{equation}
\resizebox{1.0\columnwidth}{!}{$
\displaystyle
\mathcal{L}_{\text{MWD}} = \mathbb{E}_{t, \mathcal{B}, \tilde{\mathbf{y}}^{2B}_{t}} \sum_{i \in \mathcal{B}} \text{JSD} \left( p_{\phi_{32}}^{8B,(i)}(\cdot \mid x, \tilde{\mathbf{y}}^{2B}_{t}, t) \parallel p_{\theta_B}^{2B,(i)}(\cdot \mid x, \tilde{\mathbf{y}}^{2B}_{t}, t) \right)
$}
\end{equation}
This stage, unlike the previous two stages, 
is not mandatory. 
It enables the compact diffusion model to inherit the complex perception, 
understanding, 
and planning capabilities of the 8B foundation while maintaining efficient parallel decoding.

\subsection{Model Architecture and Training}
\label{sec:network}

\paragraph{Model Architecture.}
WAM-Diff2 builds upon the Qwen3-VL framework~\cite{bai2025qwen3}, 
comprising a visual encoder, a cross-modal adapter, and a decoder-only Transformer backbone. We instantiate our framework at two scales:
\textit{(1) 8B Variant}:
Integrates a SigLIP2-SO-400M visual encoder (27 blocks, 4096 hidden dimensions)~\cite{tschannen2025siglip} with 36 Transformer blocks.
\textit{(2) 2B Variant}: Utilizes a SigLIP2-Large encoder (24 blocks, 2048 hidden dimensions) paired with 28 Transformer blocks.
To accommodate arbitrary camera aspect ratios, 
we apply a dynamic resolution strategy using 2D RoPE~\cite{su2024roformer} to interpolate spatial coordinates. 
All modalities—including linguistic tokens, 
2D bounding boxes, 
and future waypoints—are processed through a unified text tokenizer, 
entirely removing task-specific projection heads. 
Crucially, the standard causal attention mask is modified into a block-causal pattern: 
tokens within the same decoding block employ bidirectional parallel attention, while inter-block dependencies retain causal constraints.

\paragraph{System-Level Inference Optimizations.}
To fully exploit the parallel decoding capability of the discrete diffusion formulation, 
we implement two hardware-level optimizations.
\textit{FlashInfer}~\cite{ye2025flashinfer}:
We deploy customized attention kernels optimized for the block-causal attention pattern, 
accelerating intra-block bidirectional token refinement by maximizing shared memory throughput.
\textit{CUDA Graphs}~\cite{ghosh2025pygraph}:
Given that parallel remasking operates over a static number of denoising steps $T$ with fixed tensor shapes, 
we encapsulate the entire execution graph. 
This eliminates CPU launch overheads, boosting decoding throughput.

\begin{table*}[!t]
\centering
\caption{
Motion planning performance on NAVSIM v1 and v2 benchmarks.
$^\dagger$: Methods using reinforcement learning or reinforcement fine-tuning; 
$^*$: Results obtained via score-based candidate selection~\cite{yao2026drivesuprim}. 
NAVSIM v2 results utilize the original evaluation protocol for consistency. 
}
\label{tab:navsim_v1_v2_results}

\setlength{\tabcolsep}{2.1pt}
\renewcommand{\arraystretch}{0.9}
\scriptsize

\resizebox{\textwidth}{!}{
\begin{tabular}{
    @{}l
    ccccc
    >{\columncolor{gray!12}}c
    @{\hspace{5pt}}
    ccccccccc
    >{\columncolor{gray!12}}c
    @{}
}
\toprule

\textbf{Method}
& \multicolumn{6}{c}{\textbf{NAVSIM v1}}
& \multicolumn{10}{c}{\textbf{NAVSIM v2}} \\

\cmidrule(lr){2-7}
\cmidrule(lr){8-17}

& \textbf{NC}$\uparrow$
& \textbf{DAC}$\uparrow$
& \textbf{TTC}$\uparrow$
& \textbf{Comf.}$\uparrow$
& \textbf{EP}$\uparrow$
& \textbf{PDMS}$\uparrow$

& \textbf{NC}$\uparrow$
& \textbf{DAC}$\uparrow$
& \textbf{DDC}$\uparrow$
& \textbf{TLC}$\uparrow$
& \textbf{EP}$\uparrow$
& \textbf{TTC}$\uparrow$
& \textbf{LK}$\uparrow$
& \textbf{HC}$\uparrow$
& \textbf{EC}$\uparrow$
& \textbf{EPDMS}$\uparrow$ \\

\midrule

TransFuser~\cite{transfuser}
& 97.7 & 92.8 & 92.8 & 100.0 & 79.2 & 84.0
& 96.9 & 89.9 & 97.8 & 99.7 & 87.1
& 95.4 & 92.7 & 98.3 & 87.2 & 76.7 \\

Hydra-MDP++~\cite{hydramdp}
& 97.6 & 96.0 & 93.1 & 100.0 & 80.4 & 86.6
& 97.2 & 97.5 & 99.4 & 99.6 & 83.1
& 96.5 & 94.4 & 98.2 & 70.9 & 81.4 \\

ARTEMIS~\cite{feng2025artemis}
& 98.3 & 95.1 & 94.3 & 100.0 & 81.4 & 87.0
& 98.3 & 95.1 & 98.6 & 99.8 & 81.5
& 97.4 & 96.5 & 98.3 & -- & 83.1 \\

DriveSuprim~\cite{yao2026drivesuprim}
& 97.8 & 97.3 & 93.6 & 100.0 & 86.7 & 89.9
& 97.5 & 96.5 & 99.4 & 99.6 & 88.4
& 96.6 & 95.5 & 98.3 & 77.0 & 83.1 \\

DiffusionDrive~\cite{liao2025diffusiondrive}
& 98.2 & 96.2 & 94.7 & 100.0 & 82.2 & 88.1
& 98.2 & 95.9 & 99.4 & 99.8 & 87.5
& 97.3 & 96.8 & 98.3 & 87.7 & 84.5 \\

ReCogDrive$^\dagger$~\cite{li2025recogdrive}
& 97.9 & 97.3 & 94.9 & 100.0 & 87.3 & 90.8
& 98.3 & 95.2 & 99.5 & 99.8 & 87.1
& 97.5 & 96.6 & 98.3 & 86.5 & 83.6 \\

SGDrive-2B~\cite{li2026sgdrive}
& 98.6 & 95.1 & 95.4 & 100.0 & 81.2 & 87.4
& 98.6 & 94.3 & 99.5 & 99.9 & 86.0
& 97.9 & 96.1 & 98.3 & 85.9 & 86.2 \\

DriveVLA-W0~\cite{li2025drivevlaw0}
& 98.7 & 99.1 & 95.3 & 99.3 & 83.3 & 90.2
& 99.0 & 98.4 & 99.3 & 99.9 & 87.0
& 98.1 & 93.2 & 97.9 & 58.9 & 86.5 \\

ResAD~\cite{zheng2026resad}
& 98.0 & 97.5 & 94.1 & 100.0 & 83.3 & 88.8
& 97.8 & 97.2 & 99.5 & 99.8 & 88.2
& 96.9 & 97.0 & 98.4 & 88.2 & 85.5 \\

DriveFine$^\dagger$~\cite{dang2026drivefine}
& 98.6 & 97.9 & 95.2 & 99.9 & 85.5 & 90.7
& 98.7 & 97.3 & 98.8 & 99.8 & 88.2
& 97.8 & 97.7 & 98.4 & 84.7 & 87.1 \\

DiffusionDriveV2$^\dagger$~\cite{diffusiondrivev2}
& 98.3 & 97.9 & 94.8 & 99.9 & 87.5
& \textbf{91.2}
& 97.7 & 96.6 & 99.2 & 99.8 & 88.9
& 97.2 & 96.0 & 97.8 & 91.0 & 85.5 \\

\midrule

\textbf{Ours ($B=1$ AR Baseline)}
& 98.4 & 96.2 & 94.8 & 99.9 & 82.4 & 88.1
& 98.4 & 96.2 & 99.5 & 99.8 & 87.5
& 97.6 & 97.5 & 98.2 & 86.1 & 88.3 \\

\textbf{Ours ($B=32$)}
& 98.6 & 96.1 & 95.5 & 99.8 & 82.1 & 88.3
& 98.6 & 96.1 & 99.4 & 99.9 & 87.0
& 97.9 & 97.6 & 98.2 & 87.7
& \underline{88.6} \\

\textbf{Ours ($B=32$)$^*$}
& 99.1 & 98.0 & 96.7 & 99.9 & 85.1
& \underline{91.1}
& 99.1 & 98.0 & 99.4 & 99.8 & 88.0
& 98.5 & 97.2 & 98.2 & 86.4
& \textbf{90.7} \\

\bottomrule
\end{tabular}
}
\end{table*}

\paragraph{Staged Training.}
The transition from an autoregressive generalist to a parallel diffusion agent proceeds through a unified multi-task pipeline spanning two key datasets (DriveLM~\cite{sima2024drivelm}, LingoQA~\cite{marcu2024lingoqa}for scene reasoning; 
COCO~\cite{lin2014microsoft} for object detection; 
NAVSIM~\cite{dauner2024navsim} and Bench2Drive~\cite{jia2024bench2drive} for motion planning).

\textit{Phase 1: Multi-Task Autoregressive Pretraining (Optional).} 
The base models are trained for 5 epochs using standard causal next-token prediction to establish a robust multi-task semantic foundation across the consolidated training mixture. 
If a pre-trained autoregressive VLA generalist is available, this phase can be entirely bypassed, 
allowing the framework to directly initialize from the existing model weights.

\textit{Phase 2: Progressive Adaptation \& Block-Wise Distillation.}
During this phase, we gradually relax the causal constraints by expanding the block size $B \in \{4, 8, 16, 32\}$, fine-tuning for five epochs.
Each variant is initialized from the weights of its predecessor ($\theta_{B}^{(0)} \leftarrow \theta_{B/2}$). 
Concurrently, the larger-block $B_{student}$ students are distilled against the stable $B=B_{student}/2$ diffusion teacher using a symmetric JSD loss over intermediate noisy states, directly mitigating exposure bias.
Specifically, we train the larger-block student variants for five epochs per adaptation step.

\textit{Phase 3: Model-Wise Cross-Scale Distillation.}
To close the capacity gap, the 2B student ($B=32$) is distilled for 5 epochs against the advanced 8B diffusion teacher ($B=32$). 
By minimizing the JSD between their predictive token distributions, 
the compact student effectively inherits the complex cognitive reasoning of the larger foundation model without sacrificing parallel inference throughput.

\section{Experiments}
\label{sec:experiment}

\subsection{Experimental Setups}
\label{sec:exp_setup}

\paragraph{Implementation Details.} 
The architecture is built upon the Qwen3-VL~\cite{bai2025qwen3} foundation, with multi-view camera inputs resized to $1920 \times 1080$ pixels. 
All optimization phases utilize the AdamW optimizer with a weight decay of 0.05, a cosine learning rate scheduler, and a 10\% linear warmup under a global batch size of 128 across 32 ASCEND 910C NPUs.
Autoregressive pretraining runs for 5 epochs with a base learning rate of $4 \times 10^{-5}$. 
During the block-causal adaptation and block-wise distillation stages, the backbone learning rate is maintained at $4 \times 10^{-5}$ while the visual encoder uses a decoupled rate of $2 \times 10^{-6}$ to stabilize cross-modal alignment. 
The final model-wise cross-scale distillation applies a unified learning rate of $2 \times 10^{-6}$ to both components.

\paragraph{Evaluation Benchmarks.} 
We evaluate WAM-Diff2 across five benchmarks covering motion planning, scene understanding, and visual perception. 
Open-loop and closed-loop trajectory planning are assessed via NAVSIM~\cite{dauner2024navsim} (v1/v2) and Bench2Drive~\cite{jia2024bench2drive}, reporting the standard and extended Predictive Driver Model Score (PDMS and EPDMS), Success Rate (SR), and Driving Score (DS). 
Multimodal driving scene reasoning is quantified on LingoQA~\cite{marcu2024lingoqa} and DriveBench~\cite{xie2025vlms} using Lingo-Judge and GPT Score, respectively. 
General spatial grounding performance is measured via mean Average Precision (mAP) on COCO~\cite{lin2014microsoft}.

\paragraph{Evaluation Protocols.} 
To ensure rigorous comparison against both multi-task foundations and specialized planners, we establish two evaluation paradigms. 
The \textit{unified multi-task protocol} evaluates a single frozen checkpoint simultaneously across all perception, understanding, and planning tasks to verify cross-task capability preservation under the autoregressive-to-diffusion transition. 
Conversely, the \textit{task-specific protocol} trains and evaluates the architecture exclusively on isolated navigation benchmarks, ensuring fair comparisons against specialized, planning-only baselines.

\begin{table*}[!t]
\centering
\caption{
Closed-loop evaluation and multi-ability assessment on the Bench2Drive benchmark. 
We report the standard driving metrics (Efficiency, Comfort, Success Rate, and Driving Score) alongside task-specific success rates (\%) across safety-critical conditions to evaluate robustness against distribution shifts. 
\textbf{Bold} and \underline{underline} denote the best and second-best results.
}
\label{tab:bench2drive}
\footnotesize
\setlength{\tabcolsep}{1.4pt}
\renewcommand{\arraystretch}{1}

\begin{tabular*}{\textwidth}{@{\extracolsep{\fill}}lccc>{\columncolor{gray!12}}cccccc>{\columncolor{gray!12}}c@{}}
\toprule
\multirow{2}{*}{\textbf{Method}} 
& \multicolumn{4}{c}{\textbf{Closed-loop Metric} $\uparrow$}
& \multicolumn{6}{c}{\textbf{Multi-Ability Test (\%)} $\uparrow$} \\
\cmidrule(lr){2-5}
\cmidrule(lr){6-11}
& \textbf{Eff.} 
& \textbf{Comf.} 
& \textbf{Succ.} 
& \textbf{DS} 
& \textbf{Merge} 
& \textbf{Overtake} 
& \makecell[c]{\textbf{Emerg.}\\\textbf{Brake}} 
& \makecell[c]{\textbf{Give}\\\textbf{Way}} 
& \makecell[c]{\textbf{Traf.}\\\textbf{Sign}} 
& \textbf{Mean} \\
\midrule

        TCP*~\cite{wu2022trajectory}
        & 54.26 & \underline{47.80} & 15.00 & 40.70 
        & 16.18 & 20.00 & 20.00 & 10.00 & 6.99 & 14.63 \\

        TCP-ctrl*~\cite{wu2022trajectory}
        & 55.97 & \textbf{51.51} & 7.27 & 30.47 
        & 10.29 & 4.44 & 10.00 & 10.00 & 6.45 & 8.23 \\

        TCP-traj*~\cite{wu2022trajectory}
        & 76.54 & 18.08 & 30.00 & 59.90 
        & 8.89 & 24.29 & 51.67 & \underline{40.00} & 46.28 & 34.22 \\

        ThinkTwice~\cite{jia2023think}
        & 76.93 & 16.22 & 3.13 & 62.44 
        & 27.38 & 18.42 & 35.82 & \textbf{50.00} & 54.23 & 37.17 \\

        DriveAdapter*~\cite{jia2023driveadapter}
        & 70.22 & 16.01 & 33.08 & 64.22 
        & 28.82 & 26.38 & 48.76 & \textbf{50.00} & 56.43 & 42.08 \\

        AD-MLP~\cite{zhai2023rethinking}
        & 48.45 & 22.63 & 0.00 & 18.05 
        & 0.00 & 0.00 & 0.00 & 0.00 & 4.35 & 0.87 \\

        UniAD-T.~\cite{hu2023planning}
        & 123.92 & 47.04 & 13.18 & 40.73 
        & 8.89 & 9.33 & 20.00 & 20.00 & 15.43 & 14.73 \\

        UniAD-B.~\cite{hu2023planning}
        & 129.21 & 43.58 & 16.36 & 45.81 
        & 14.10 & 17.78 & 21.67 & 10.00 & 14.21 & 15.55 \\

        VAD~\cite{jiang2023vad}
        & 157.94 & 46.01 & 15.00 & 42.35 
        & 8.11 & 24.44 & 18.64 & 20.00 & 19.15 & 18.07 \\

        ReCogDrive~\cite{li2025recogdrive} 
        & 138.18 & 17.45 & 45.45 & 71.36 
        & 29.73 & 20.00 & \underline{69.09} & 20.00 & 71.34 & 42.03 \\

        Orion~\cite{fu2025orion}
        & 151.48 & 17.38 & \textbf{54.62} & 77.74 
        & 25.00 & \underline{71.11} & \textbf{78.33} & 33.00 & 69.15 & 54.72 \\
        
        DriveMoE~\cite{yang2026drivemoe}
        & \underline{175.96} & 15.31 & 48.64 & 74.22 
        & 34.67 & 40.00 & 65.45 & \underline{40.00} & 59.44 & 47.91 \\

        UniDriveVLA~\cite{li2026unidrivevla}
        & \textbf{198.86} & 11.78 & 51.82 & 78.37 
        & 38.75 & \textbf{80.00} & 50.00 & 30.00 & 58.95 & 51.53 \\
        
        \midrule
        \textbf{Ours-2B (B=1, AR Baseline)} & 118.10 & 21.40 & \underline{51.96} & \textbf{80.51} & 45.59 & 62.79 & 60.34 & \textbf{50.00} & \textbf{89.25} & \textbf{61.59} \\
        
        \textbf{Ours-2B (B=4)} & 116.24 & 22.01 & 51.36 & \underline{80.01} & \textbf{47.50} & 60.00 & 60.00 & \textbf{50.00} & \underline{88.42} & \underline{61.18} \\
        
        \textbf{Ours-2B (B=8)} & 118.17 & 19.26 & 50.23 & 79.73 & \underline{46.25} & 62.22 & 60.00 & \textbf{50.00} & 87.37 & 61.17 \\
        
        \textbf{Ours-2B (B=16)} & 110.45 & 21.36 & 50.45 & 79.52 & 45.00 & 60.00 & 61.67 & \textbf{50.00} & 87.37 & 60.81 \\
        
        \textbf{Ours-2B (B=32)} & 108.20 & 23.15 & 49.55 & 78.93 & 43.75 & 62.22 & 61.67 & \textbf{50.00} & 85.26 & 60.58 \\
        \bottomrule
\vspace{-5mm}
\end{tabular*}
\end{table*}

\subsection{Comparison with Existing Works}
We evaluate WAM-Diff2 under two evaluation paradigms: 
a \textit{unified multi-task protocol} (Table~\ref{tab:main_results}) that assesses multi-capability preservation and inference throughput using a single model, 
and a \textit{task-specific protocol} (Tables~\ref{tab:navsim_v1_v2_results} and~\ref{tab:bench2drive}) for equitable comparison against specialized driving planners.

\paragraph{Unified Multi-Task Performance.}
As shown in Table~\ref{tab:main_results}, 
our 2B diffusion model with a block size of $B=32$ achieves a DriveBench score of 48.80 and a LingoQA metric of 65.80, 
matching or exceeding the semantic capabilities of mature autoregressive generalists sharing the same backbone. 
For spatial perception, 
the model retains strong visual perception with 36.3 mAP on COCO, 
showing that dense cross-modal alignment remains intact through the generative paradigm transition. 
Furthermore, without relying on auxiliary reinforcement fine-tuning or score-based candidate selection, 
WAM-Diff2 delivers a robust open-loop planning performance of 87.44 PDMS on NAVSIM v1. 
These results demonstrate that transitioning to parallel discrete diffusion successfully maintains holistic scene reasoning and planning precision while unlocking substantial gains in decoding efficiency.

\begin{figure*}[t]
    \centering
    \includegraphics[width=\linewidth]{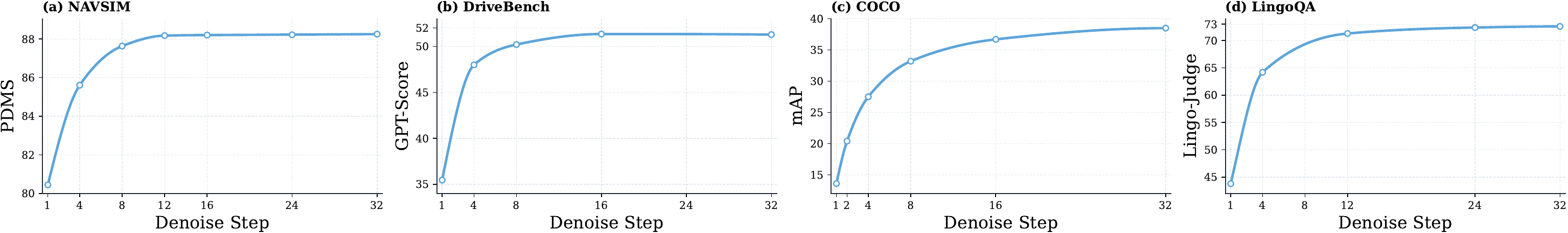}
    \caption{
    Effect of denoising steps.
    PDMS improves as the number of denoising steps increases and
    gradually saturates at approximately 8--16 steps.
    }
    \label{fig:denoising_steps}
    \vspace{-3mm}
\end{figure*}
\begin{figure*}[t]
    \centering
    \includegraphics[width=\linewidth]{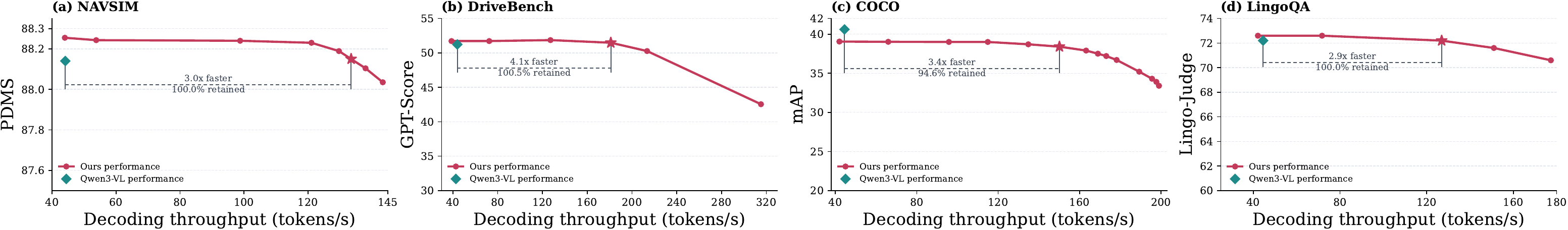}
    \caption{{Performance--efficiency trade-offs of WAM-Diff2 across
    multiple benchmarks.}
    WAM-Diff2 consistently preserves driving understanding,
    general visual perception, and motion-planning performance
    while achieving higher decoding throughput.}
    \label{fig:tradeoff}
\end{figure*}

\begin{figure}[t]
    \centering
    \includegraphics[width=\linewidth]{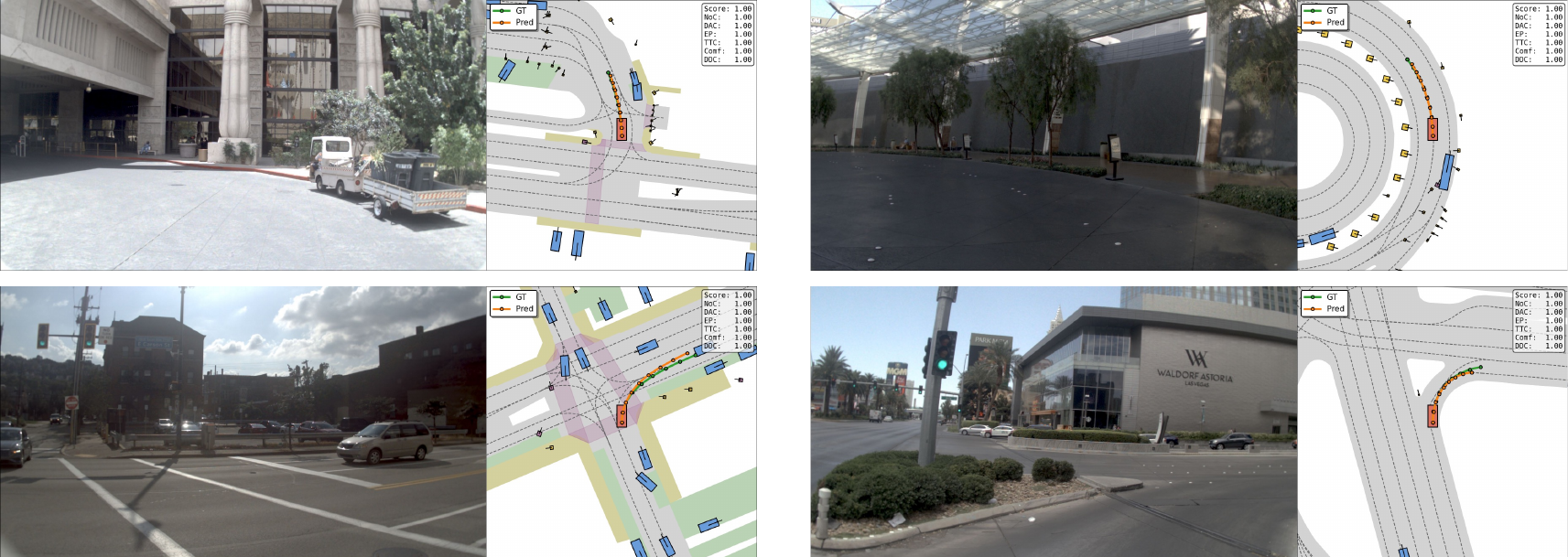}
    \vspace{-3mm}
    \caption{
    Qualitative results of WAM-Diff2 on NAVSIM. The predicted trajectories demonstrate robust planning behavior across diverse driving scenarios.
    }
    \label{fig:navsim_vis}
    \vspace{-5mm}
\end{figure}

\begin{figure}[t]
    \centering
    \includegraphics[width=\linewidth]{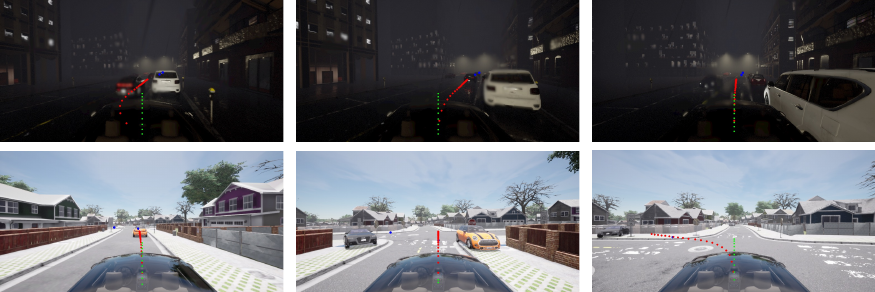}
    \vspace{-3mm}
    \caption{
    Qualitative results of WAM-Diff2 on Bench2Drive. The predicted trajectories demonstrate robust planning behavior across diverse driving scenarios.
    }
    \label{fig:bench2drive_vis}
    \vspace{-5mm}
\end{figure}

\paragraph{Task-Specific Motion Planning.} 
Under the planning isolated evaluation framework, WAM-Diff2 preserves the planning capability of its autoregressive counterpart, demonstrating that the conversion from autoregressive to diffusion decoding introduces no meaningful performance degradation. 
On the open-loop NAVSIM benchmark (Table~\ref{tab:navsim_v1_v2_results}), 
our model matches or surpasses the original autoregressive baseline, 
achieving 88.3 PDMS (v1) and 88.6 EPDMS (v2). 
When coupled with score-based candidate selection ($^*$), 
WAM-Diff2 reaches a peak performance of 91.1 PDMS and 90.7 EPDMS, 
outperforming advanced specialized driving agents such as ReCogDrive~\cite{li2025recogdrive} and DriveVLA-W0~\cite{li2025drivevlaw0}. 
In closed-loop evaluations on Bench2Drive (Table~\ref{tab:bench2drive}), 
WAM-Diff2 demonstrates robust generalization under extreme distribution shifts, 
attaining a Success Rate of 
49.55\% 
and a Driving Score of 78.93. 
Crucially, it yields pronounced improvements in safety-critical scenarios like Emergency Braking (61.67\%) 
and Traffic Sign compliance 
(85.26\%), 
underscoring the benefits of our distillation framework in mitigating inference-time exposure bias. Corresponding trajectory rollouts are visualized in Figures~\ref{fig:navsim_vis} and~\ref{fig:bench2drive_vis}.

\paragraph{Block-Size Scaling and Efficiency.} 
Table~\ref{tab:main_results} further quantifies the latency-accuracy Pareto frontier across varied decoding block sizes $B$. 
Expanding $B$ from 4 to 32 yields substantial hardware parallelization: 
vanilla implementation throughput rises from 68.3 to 124.8 TPS, 
which scales dramatically to a peak of 673.4 TPS when integrated with system-level FlashInfer and CUDA Graph optimizations. 
This represents a cumulative $15.1\times$ inference acceleration over the sequential autoregressive baseline. 
Importantly, this throughput optimization introduces negligible performance degradation: 
motion-planning metrics remain remarkably stable ($\Delta \le 0.5$ points on NAVSIM v1/v2), 
while open-ended visual question answering and grounding exhibit only minor trade-offs.

\begin{table*}[!t]
\centering
\captionsetup{font=footnotesize}

\begin{minipage}[t]{0.52\textwidth}
    \vspace{0pt}
    \centering

    \caption{
        Cumulative ablation of AR-to-diffusion VLA distillation.
        Starting from the AR baseline, we compare direct AR-to-diffusion
        adaptation, block-wise distillation, and model-wise distillation.
        All diffusion variants are evaluated under the same decoding
        configuration on NAVSIM.
    }
    \label{tab:stage_ablation}

    \vspace{-2mm}

    {\ablationtablesetup
    \begin{tabularx}{\linewidth}{@{}L*{6}{c}@{}}
        \toprule
        \textbf{Setting}
        & \textbf{NC}$\uparrow$
        & \textbf{DAC}$\uparrow$
        & \textbf{TTC}$\uparrow$
        & \textbf{Comf.}$\uparrow$
        & \textbf{EP}$\uparrow$
        & \textbf{PDMS}$\uparrow$ \\
        \midrule

        AR baseline
        & 98.4
        & \textbf{96.2}
        & 94.8
        & \textbf{99.9}
        & \textbf{82.4}
        & 88.1 \\

        + Direct AR-to-diffusion adaptation
        & 97.9
        & 93.0
        & 93.2
        & 99.3
        & 78.5
        & 84.1 \\

        + Block-wise distillation
        & 98.2
        & 95.7
        & 95.0
        & 99.3
        & 81.6
        & 87.7 \\

        + Model-wise distillation
        & \textbf{98.6}
        & 96.1
        & \textbf{95.5}
        & 99.8
        & 82.1
        & \textbf{88.3} \\

        \bottomrule
    \end{tabularx}
    }

    \vspace{-3mm}
\end{minipage}
\hfill
\begin{minipage}[t]{0.46\textwidth}
    \vspace{0pt}
    \centering

    \caption{
        Ablation of divergence objectives for AR-to-diffusion VLA
        distillation. Specifically, we compare forward KL, reverse KL, and
        Jensen--Shannon divergence (JSD) for motion planning on the
        NAVSIM benchmark, while keeping all other training and evaluation
        settings fixed.
    }
    \label{tab:distill_loss}

    \vspace{-2.2mm}

    {\ablationtablesetup
    \begin{tabularx}{\linewidth}{@{}L*{6}{c}@{}}
        \toprule
        \textbf{Distillation Loss}
        & \textbf{NC}$\uparrow$
        & \textbf{DAC}$\uparrow$
        & \textbf{TTC}$\uparrow$
        & \textbf{Comf.}$\uparrow$
        & \textbf{EP}$\uparrow$
        & \textbf{PDMS}$\uparrow$ \\
        \midrule

        Forward KL
        & \textbf{98.6}
        & 96.0
        & \textbf{95.6}
        & \textbf{99.9}
        & 81.9
        & 88.2 \\

        Reverse KL
        & 98.5
        & 95.9
        & 95.3
        & 99.7
        & 81.8
        & 88.0 \\

        JSD
        & \textbf{98.6}
        & \textbf{96.1}
        & 95.5
        & 99.8
        & \textbf{82.1}
        & \textbf{88.3} \\

        \bottomrule
    \end{tabularx}
    }

    \vspace{-7mm}
\end{minipage}

\end{table*}

\begin{table*}[!t]
\centering
\captionsetup{font=footnotesize}

\begin{minipage}[t]{0.48\textwidth}
    \vspace{0pt}
    \centering

    \caption{
        Top-$K$ overlap ratio between different teachers and the
        diffusion student on training samples.
        All values are reported as percentages.
        Higher $\rho_K$ indicates stronger teacher--student alignment.
    }
    \label{tab:teacher_student_topk}

    \vspace{-2mm}

    {\ablationtablesetup
    \begin{tabularx}{\linewidth}{@{}L*{4}{c}@{}}
        \toprule
        \textbf{Teacher}
        & $\boldsymbol{\rho_1}\uparrow$
        & $\boldsymbol{\rho_5}\uparrow$
        & $\boldsymbol{\rho_{10}}\uparrow$
        & $\boldsymbol{\rho_{20}}\uparrow$ \\
        \midrule

        AR Teacher
        & 83.6
        & 51.2
        & 54.3
        & 46.1 \\

        Diffusion Teacher
        & \textbf{84.8}
        & \textbf{58.2}
        & \textbf{60.5}
        & \textbf{50.7} \\

        \bottomrule
    \end{tabularx}
    }

    \vspace{-3mm}
\end{minipage}
\hfill
\begin{minipage}[t]{0.5\textwidth}
    \vspace{0pt}
    \centering

    \caption{
        Ablation of different teacher models for motion planning on the
        NAVSIM benchmark. We compare autoregressive teacher and diffusion
        teacher under otherwise identical distillation settings.
    }
    \label{tab:teacher_ablation}

    \vspace{-2.1mm}

    {\ablationtablesetup
    \begin{tabularx}{\linewidth}{@{}L*{6}{c}@{}}
        \toprule
        \textbf{Teacher Model}
        & \textbf{NC}$\uparrow$
        & \textbf{DAC}$\uparrow$
        & \textbf{TTC}$\uparrow$
        & \textbf{Comf.}$\uparrow$
        & \textbf{EP}$\uparrow$
        & \textbf{PDMS}$\uparrow$ \\
        \midrule

        AR Teacher
        & 97.2
        & 89.5
        & 91.7
        & 97.3
        & 74.6
        & 79.9 \\

        Diffusion Teacher
        & \textbf{98.6}
        & \textbf{96.1}
        & \textbf{95.5}
        & \textbf{99.8}
        & \textbf{82.1}
        & \textbf{88.3} \\

        \bottomrule
    \end{tabularx}
    }

    \vspace{-7mm}
\end{minipage}

\end{table*}

\begin{figure}[t]
    \centering
    \includegraphics[width=\linewidth]{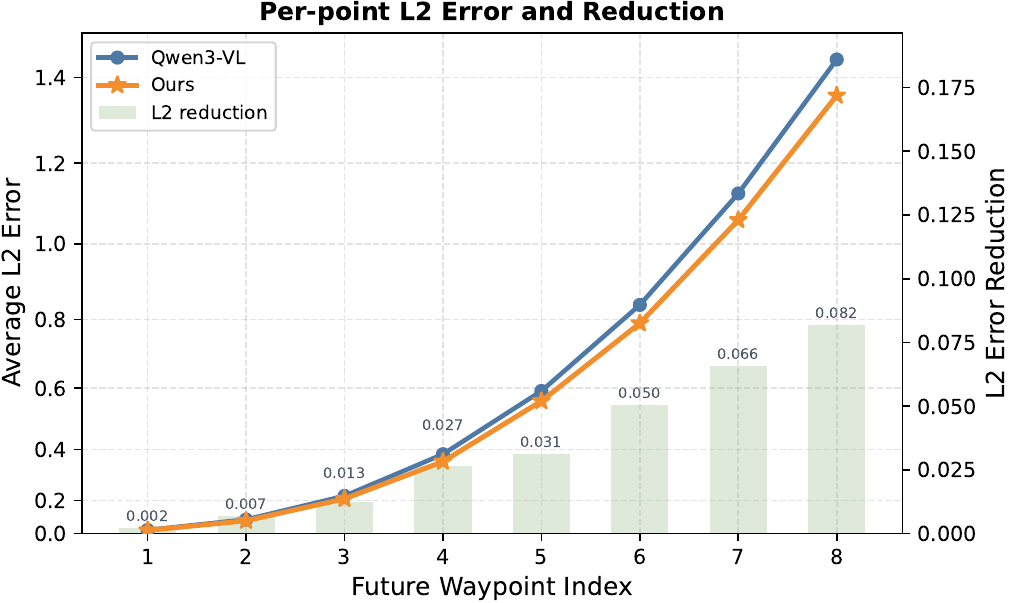}

    \vspace{-1mm}
    \caption{
    Per-waypoint L2 error under inference on the NAVSIM dataset.
    The L2 errors of the autoregressive baseline gradually increase compared to our method, and there is a trend of error amplification.
    These results suggest that bidirectional diffusion generation help to alleviate error accumulation caused by autoregressive exposure bias.
    }
    \label{fig:long_horizon_error_accumulation}
    \vspace{-2mm}
\end{figure}

\subsection{Ablation and Discussion}
\paragraph{Effectiveness of AR-to-Diffusion Transformation Stages.}
We quantify the contribution of each stage in our framework using the NAVSIM benchmark~\cite{dauner2024navsim} (Table \ref{tab:stage_ablation}). 
Specifically, direct AR-to-diffusion adaptation results in a significant performance drop (84.1 PDMS) due to the radical shift from causal to bidirectional attention. 
Block-wise distillation recovers substantial performance (87.7) by aligning the student with a stable, small-block diffusion teacher. 
Model-wise distillation further closes the gap, 
allowing the 2B student to inherit high-level reasoning from the 8B teacher, 
ultimately achieving 88.3 PDMS.

\paragraph{Paradigm-Consistent Distillation Mechanics.}
To optimize knowledge transfer across generative paradigms, 
we evaluate both the teacher configuration and the token-level alignment objectives. 
As shown in Table~\ref{tab:teacher_student_topk}, 
the 8B diffusion teacher exhibits significantly higher top-$K$ token overlap ($\rho_5 = 58.2\%$) with the 2B diffusion student compared to the 8B autoregressive teacher ($\rho_5 = 51.2\%$). 
This structural alignment translates to a severe performance drop when distilling from the autoregressive teacher ($79.9$ PDMS; Table~\ref{tab:teacher_ablation}), 
showing that paradigm consistency is prerequisite for effective cross-scale distillation. 
Regarding the alignment objective (Table~\ref{tab:distill_loss}), 
while forward and reverse KL divergences yield comparable baselines, 
the symmetric JSD achieves peak performance ($88.3$ PDMS) by preserving both the mode-seeking precision and broad semantic coverage required for multi-modal trajectory distributions.

\begin{figure}[!t]
    \centering
    \includegraphics[width=1.0\linewidth]{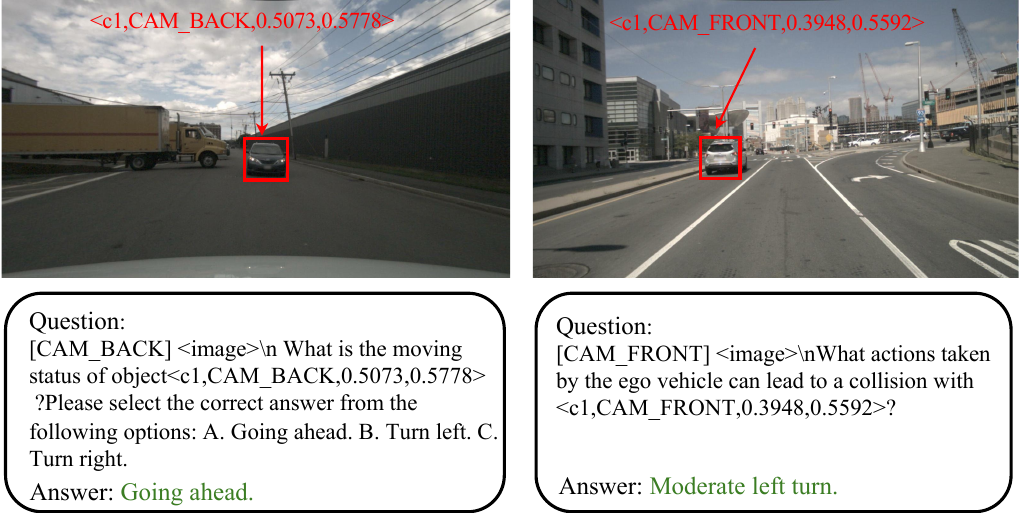}
    \vspace{-5mm}
    \caption{Visualization of understanding results. }
    \vspace{-3mm}
    \label{fig:driving_understanding}
\end{figure}

\begin{figure}[!t]
    \centering
    \includegraphics[width=1.0\linewidth]{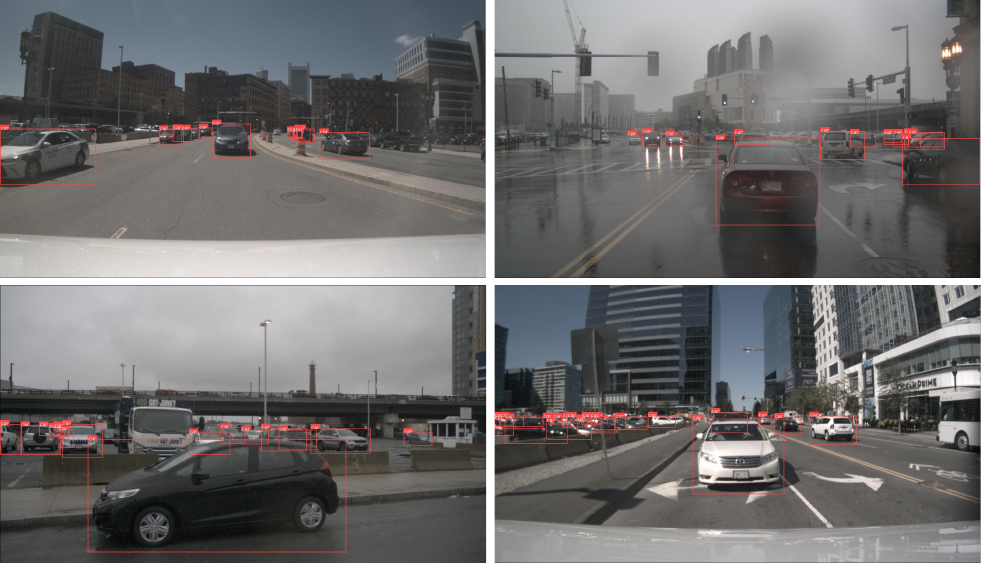}
    \vspace{-5mm}
    \caption{Visualization of perception results. }
    \vspace{-3mm}
    \label{fig:perception}
\end{figure}

\paragraph{Speed--Accuracy Pareto Frontier.}
We analyze the inference dynamics of WAM-Diff2 across varying refinement budgets and decoding block sizes ($B$). 
As illustrated in Figure~\ref{fig:denoising_steps}, 
multi-task performance scales rapidly with additional denoising steps, saturating at approximately 8--16 iterations across all benchmarks. 
Figure~\ref{fig:tradeoff} further maps the performance--efficiency trade-off: trajectory planning and driving understanding remain robust across a wide throughput range, whereas visual grounding (COCO mAP) exhibits earlier degradation due to its strict coordinate-formatting constraints. By adjusting $B$ from 4 to 32 (Table~\ref{tab:main_results}), 
vanilla decoding throughput rises from 68.3 to 124.8 TPS with negligible planning degradation ($\Delta \le 0.5$ PDMS), 
demonstrating flexible, runtime-configurable control over execution latency.

\paragraph{Efficiency Analysis.}
The throughput acceleration of WAM-Diff2 stems from unlocking parallel block-wise token refinement. 
Under a standard implementation, replacing sequential AR decoding with block-wise diffusion provides an immediate $2.8\times$ speedup on NAVSIM (Figure~\ref{fig:teaser}). 
Integrating hardware-level optimizations yields compound gains: specialized FlashInfer attention kernels maximize shared memory throughput to provide a $1.7\times$ acceleration, 
while CUDA Graphs eliminate CPU launch overheads for static tensor shapes, contributing an additional $3.1\times$ boost. 
Combined, these algorithmic and system optimizations deliver a cumulative $15.1\times$ reduction in decoding latency (from 22.7\,ms/token to 1.5\,ms/token).

\paragraph{Mitigation of Exposure Bias.}
WAM-Diff2 leverages bidirectional attention and iterative refinement to denoise tokens within each block, 
allowing the model to dynamically revise uncertain intermediate predictions rather than permanently committing to early mistakes. 
To quantify this architectural advantage, 
we evaluated WAM-Diff2 against the autoregressive baseline across 12,146 paired NAVSIM samples. 
As shown in Figure~\ref{fig:long_horizon_error_accumulation}, 
the block-wise diffusion framework reduces the average per-waypoint $L_2$ error by 5.8\% (from 0.5935 to 0.5589). 
Crucially, the absolute error reduction widens monotonically over time--from 0.002 at the initial waypoint to 0.082 at the final horizon (waypoint 8)--demonstrating that bidirectional parallel refinement effectively mitigates long-horizon error accumulation and enhances closed-loop robustness against compounding sequence drift.

\begin{figure}[!t]
    \centering
    \includegraphics[width=1.0\linewidth]{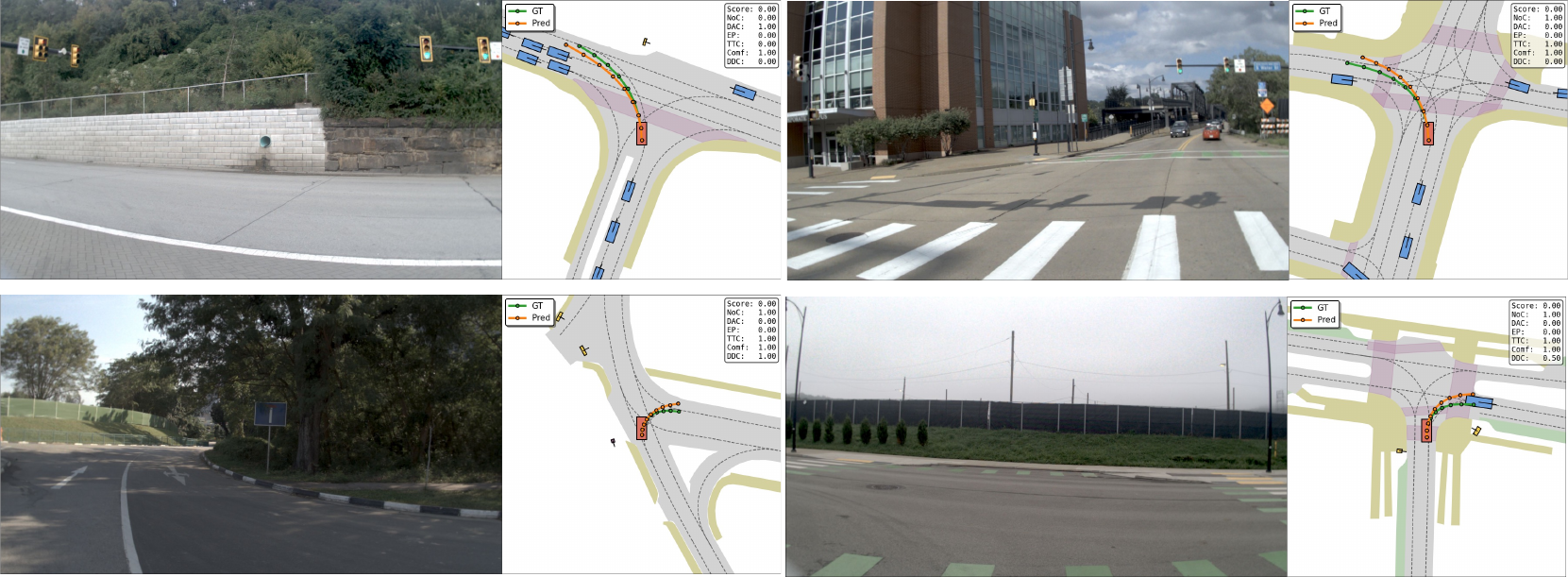}
    \vspace{-5mm}
    \caption{Failure cases visualization on NAVSIM.}
    \vspace{-3mm}
    \label{fig:navsim_failure_case}
\end{figure}

\begin{figure}[!t]
    \centering
    \includegraphics[width=1.0\linewidth]{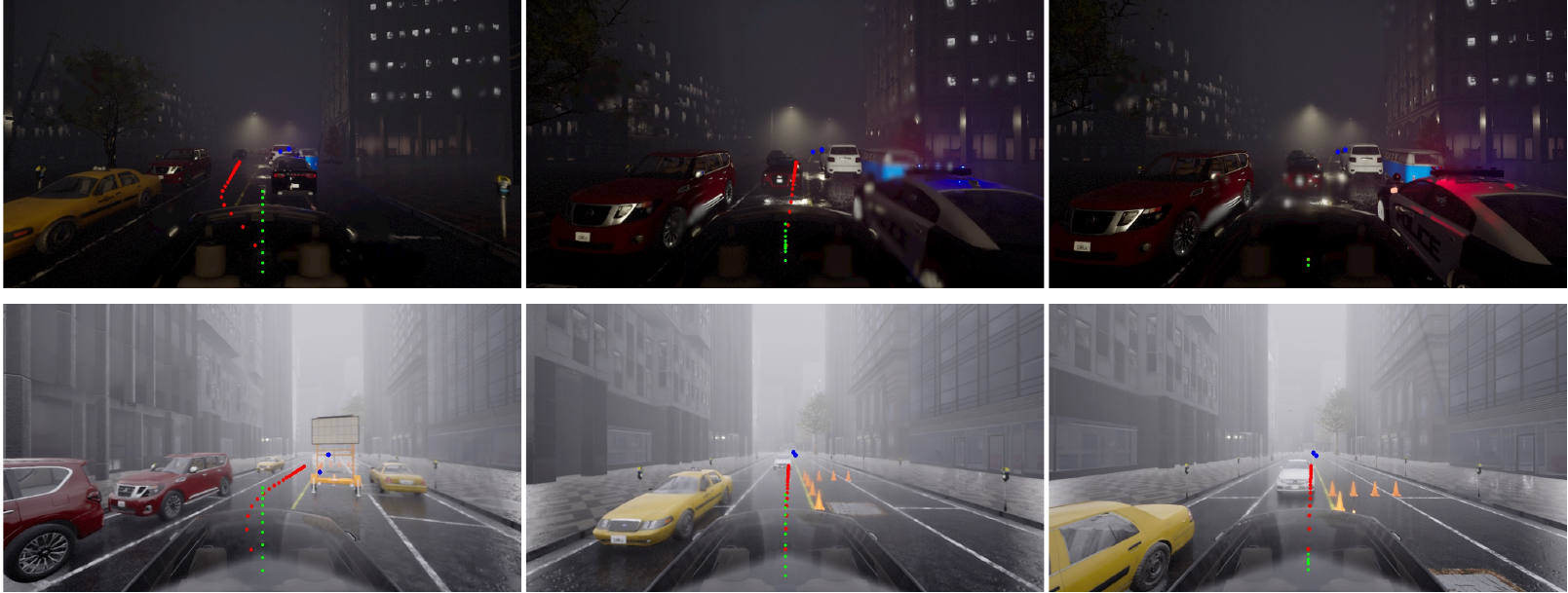}
    \vspace{-5mm}
    \caption{Failure cases visualization on Bench2Drive.}
    \vspace{-3mm}
    \label{fig:bench2drive_failure_case}
\end{figure}

\paragraph{Limitations and Failure Cases.}
Qualitative failure modes on the NAVSIM~\cite{dauner2024navsim} and Bench2Drive~\cite{jia2024bench2drive} benchmarks are visualized in Figures~\ref{fig:navsim_failure_case} and~\ref{fig:bench2drive_failure_case}. 
Despite its high inference throughput and robustness against exposure bias, 
WAM-Diff2 exhibits two primary limitations. 
First, its structural reliance on discrete tokenization can introduce spatial quantization artifacts, 
occasionally compromising the smoothness required for high-precision trajectory planning. 
Second, the downstream capabilities of the parallel diffusion student remain fundamentally upper-bounded by the baseline semantic reasoning proficiency of the initial autoregressive teacher.

\section{Conclusion}
We presented WAM-Diff2, 
a universal architecture translation paradigm that seamlessly converts pre-trained,
multi-task autoregressive VLAs into high-throughput discrete diffusion agents for autonomous driving. 
By deploying a novel three-stage hierarchical distillation strategy—spanning progressive block-wise adaptation, block-wise distillation, 
and cross-scale paradigm alignment—our framework successfully bridges the architectural chasm between sequential causal generation and parallel bidirectional refinement.
Extensive evaluations across motion planning, visual perception, and scene understanding benchmarks demonstrate that WAM-Diff2 preserves the holistic cognitive reasoning of large-scale autoregressive generalists while eliminating their structural bottlenecks. 
Algorithmically, the transition to a block-causal diffusion mechanism successfully mitigates exposure bias, 
monotonically suppressing long-horizon error accumulation during sequence rollouts. Under system-level optimizations via FlashInfer and CUDA Graphs, 
WAM-Diff2 compresses token decoding latency from 22.7 ms to 1.5 ms, 
yielding a cumulative $15.1\times$ inference acceleration. 
Ultimately, this work provides a scalable, low-cost pipeline to deploy advanced multimodal driving intelligence under the rigid throughput constraints of highly efficient autonomous systems.

\bibliography{ref}

@String(AAAI = {AAAI})

@article{hwang2024emma,
  title={Emma: End-to-end multimodal model for autonomous driving},
  author={Hwang, Jyh-Jing and Xu, Runsheng and Lin, Hubert and Hung, Wei-Chih and Ji, Jingwei and Choi, Kristy and Huang, Di and He, Tong and Covington, Paul and Sapp, Benjamin and others},
  journal={arXiv preprint arXiv:2410.23262},
  year={2024}
}

@article{fu2025orion,
  title={Orion: A holistic end-to-end autonomous driving framework by vision-language instructed action generation},
  author={Fu, Haoyu and Zhang, Diankun and Zhao, Zongchuang and Cui, Jianfeng and Liang, Dingkang and Zhang, Chong and Zhang, Dingyuan and Xie, Hongwei and Wang, Bing and Bai, Xiang},
  journal={arXiv preprint arXiv:2503.19755},
  year={2025}
}

@article{zhou2025autovla,
  title={AutoVLA: A Vision-Language-Action Model for End-to-End Autonomous Driving with Adaptive Reasoning and Reinforcement Fine-Tuning},
  author={Zhou, Zewei and Cai, Tianhui and Zhao, Seth Z and Zhang, Yun and Huang, Zhiyu and Zhou, Bolei and Ma, Jiaqi},
  journal={arXiv preprint arXiv:2506.13757},
  year={2025}
}

@article{zhou2025opendrivevla,
  title={Opendrivevla: Towards end-to-end autonomous driving with large vision language action model},
  author={Zhou, Xingcheng and Han, Xuyuan and Yang, Feng and Ma, Yunpu and Knoll, Alois C},
  journal={arXiv preprint arXiv:2503.23463},
  year={2025}
}

@inproceedings{liao2025diffusiondrive,
  title={Diffusiondrive: Truncated diffusion model for end-to-end autonomous driving},
  author={Liao, Bencheng and Chen, Shaoyu and Yin, Haoran and Jiang, Bo and Wang, Cheng and Yan, Sixu and Zhang, Xinbang and Li, Xiangyu and Zhang, Ying and Zhang, Qian and others},
  booktitle={Proceedings of the Computer Vision and Pattern Recognition Conference},
  pages={12037--12047},
  year={2025}
}

@article{xu2025wam,
  title={WAM-Diff: A Masked Diffusion VLA Framework with MoE and Online Reinforcement Learning for Autonomous Driving},
  author={Xu, Mingwang and Cui, Jiahao and Cai, Feipeng and Shang, Hanlin and Zhu, Zhihao and Luan, Shan and Xu, Yifang and Zhang, Neng and Li, Yaoyi and Cai, Jia and others},
  journal={arXiv preprint arXiv:2512.11872},
  year={2025}
}

@article{nie2025large,
  title={Large Language Diffusion Models},
  author={Nie, Shen and Zhu, Fengqi and You, Zebin and Zhang, Xiaolu and Ou, Jingyang and Hu, Jun and Zhou, Jun and Lin, Yankai and Wen, Ji-Rong and Li, Chongxuan},
  journal={arXiv preprint arXiv:2502.09992},
  year={2025}
}

@article{you2025llada,
  title={LLaDA-V: Large Language Diffusion Models with Visual Instruction Tuning},
  author={You, Zebin and Nie, Shen and Zhang, Xiaolu and Hu, Jun and Zhou, Jun and Lu, Zhiwu and Wen, Ji-Rong and Li, Chongxuan},
  journal={arXiv preprint arXiv:2505.16933},
  year={2025}
}

@article{xu2024drivegpt4,
  title={Drivegpt4: Interpretable end-to-end autonomous driving via large language model},
  author={Xu, Zhenhua and Zhang, Yujia and Xie, Enze and Zhao, Zhen and Guo, Yong and Wong, Kwan-Yee K and Li, Zhenguo and Zhao, Hengshuang},
  journal={IEEE Robotics and Automation Letters},
  volume={9},
  number={10},
  pages={8186--8193},
  year={2024},
  publisher={IEEE}
}

@article{tian2024drivevlm,
  title={Drivevlm: The convergence of autonomous driving and large vision-language models},
  author={Tian, Xiaoyu and Gu, Junru and Li, Bailin and Liu, Yicheng and Wang, Yang and Zhao, Zhiyong and Zhan, Kun and Jia, Peng and Lang, Xianpeng and Zhao, Hang},
  journal={arXiv preprint arXiv:2402.12289},
  year={2024}
}

@inproceedings{sima2024drivelm,
  title={Drivelm: Driving with graph visual question answering},
  author={Sima, Chonghao and Renz, Katrin and Chitta, Kashyap and Chen, Li and Zhang, Hanxue and Xie, Chengen and Bei{\ss}wenger, Jens and Luo, Ping and Geiger, Andreas and Li, Hongyang},
  booktitle={European conference on computer vision},
  pages={256--274},
  year={2024},
  organization={Springer}
}

@article{han2025percept,
  title={Percept-WAM: Perception-enhanced world-awareness-action model for robust end-to-end autonomous driving},
  author={Han, Jianhua and Tian, Meng and Zhu, Jiangtong and He, Fan and Zhang, Huixin and Guo, Sitong and Zhu, Dechang and Tang, Hao and Xu, Pei and Guo, Yuze and others},
  journal={arXiv preprint arXiv:2511.19221},
  year={2025}
}

@article{yang2025drivemoe,
  title={DriveMoE: Mixture-of-experts for vision-language-action model in end-to-end autonomous driving},
  author={Yang, Zhenjie and Chai, Yilin and Jia, Xiaosong and Li, Qifeng and Shao, Yuqian and Zhu, Xuekai and Su, Haisheng and Yan, Junchi},
  journal={arXiv preprint arXiv:2505.16278},
  year={2025}
}

@article{li2026unidrivevla,
  title={UniDriveVLA: Unifying Understanding, Perception, and Action Planning for Autonomous Driving},
  author={Li, Yongkang and Zhou, Lijun and Yan, Sixu and Liao, Bencheng and Yan, Tianyi and Xiong, Kaixin and Chen, Long and Xie, Hongwei and Wang, Bing and Chen, Guang and others},
  journal={arXiv preprint arXiv:2604.02190},
  year={2026}
}

@article{austin2021structured,
  title={Structured denoising diffusion models in discrete state-spaces},
  author={Austin, Jacob and Johnson, Daniel D and Ho, Jonathan and Tarlow, Daniel and Van Den Berg, Rianne},
  journal={Advances in neural information processing systems},
  volume={34},
  pages={17981--17993},
  year={2021}
}

@article{campbell2022continuous,
  title={A continuous time framework for discrete denoising models},
  author={Campbell, Andrew and Benton, Joe and De Bortoli, Valentin and Rainforth, Thomas and Deligiannidis, George and Doucet, Arnaud},
  journal={Advances in Neural Information Processing Systems},
  volume={35},
  pages={28266--28279},
  year={2022}
}

@article{lou2023discrete,
  title={Discrete diffusion modeling by estimating the ratios of the data distribution},
  author={Lou, Aaron and Meng, Chenlin and Ermon, Stefano},
  journal={arXiv preprint arXiv:2310.16834},
  year={2023}
}

@article{gat2024discrete,
  title={Discrete flow matching},
  author={Gat, Itai and Remez, Tal and Shaul, Neta and Kreuk, Felix and Chen, Ricky TQ and Synnaeve, Gabriel and Adi, Yossi and Lipman, Yaron},
  journal={Advances in Neural Information Processing Systems},
  volume={37},
  pages={133345--133385},
  year={2024}
}

@article{ye2025dream,
  title={Dream 7b: Diffusion large language models},
  author={Ye, Jiacheng and Xie, Zhihui and Zheng, Lin and Gao, Jiahui and Wu, Zirui and Jiang, Xin and Li, Zhenguo and Kong, Lingpeng},
  journal={arXiv preprint arXiv:2508.15487},
  year={2025}
}

@article{yang2025mmada,
  title={Mmada: Multimodal large diffusion language models},
  author={Yang, Ling and Tian, Ye and Li, Bowen and Zhang, Xinchen and Shen, Ke and Tong, Yunhai and Wang, Mengdi},
  journal={arXiv preprint arXiv:2505.15809},
  year={2025}
}

@article{cui2025vilad,
  title={Vilad: A large vision language diffusion framework for end-to-end autonomous driving},
  author={Cui, Can and Zhou, Yupeng and Peng, Juntong and Park, Sung-Yeon and Yang, Zichong and Sankaranarayanan, Prashanth and Zhang, Jiaru and Zhang, Ruqi and Wang, Ziran},
  journal={arXiv preprint arXiv:2508.12603},
  year={2025}
}

@article{li2025reflectdrive,
  title={Discrete diffusion for reflective vision-language-action models in autonomous driving},
  author={Li, Pengxiang and Zheng, Yinan and Wang, Yue and Wang, Huimin and Zhao, Hang and Liu, Jingjing and Zhan, Xianyuan and Zhan, Kun and Lang, Xianpeng},
  journal={arXiv preprint arXiv:2509.20109},
  year={2025}
}

@article{dang2026drivefine,
  title={DriveFine: Refining-Augmented Masked Diffusion VLA for Precise and Robust Driving},
  author={Dang, Chenxu and Ang, Sining and Li, Yongkang and Tian, Haochen and Wang, Jie and Li, Guang and Ye, Hangjun and Ma, Jie and Chen, Long and Wang, Yan},
  journal={arXiv preprint arXiv:2602.14577},
  year={2026}
}

@article{gu2023minillm,
  title={Minillm: Knowledge distillation of large language models},
  author={Gu, Yuxian and Dong, Li and Wei, Furu and Huang, Minlie},
  journal={arXiv preprint arXiv:2306.08543},
  year={2023}
}

@inproceedings{agarwal2024policy,
  title={On-policy distillation of language models: Learning from self-generated mistakes},
  author={Agarwal, Rishabh and Vieillard, Nino and Zhou, Yongchao and Stanczyk, Piotr and Garea, Sabela Ramos and Geist, Matthieu and Bachem, Olivier},
  booktitle={The twelfth international conference on learning representations},
  year={2024}
}

@article{li2026rethinking,
  title={Rethinking On-Policy Distillation of Large Language Models: Phenomenology, Mechanism, and Recipe},
  author={Li, Yaxuan and Zuo, Yuxin and He, Bingxiang and Zhang, Jinqian and Xiao, Chaojun and Qian, Cheng and Yu, Tianyu and Gao, Huan-ang and Yang, Wenkai and Liu, Zhiyuan and others},
  journal={arXiv preprint arXiv:2604.13016},
  year={2026}
}

@article{afsharrad2026policy,
  title={On-Policy Distillation of Language Models for Autonomous Vehicle Motion Planning},
  author={Afsharrad, Amirhossein and Abedsoltan, Amirhesam and Moradipari, Ahmadreza and Lall, Sanjay},
  journal={arXiv preprint arXiv:2604.07944},
  year={2026}
}

@inproceedings{xie2025vlms,
  title={Are VLMs Ready for Autonomous Driving? An Empirical Study from the Reliability, Data and Metric Perspectives},
  author={Xie, Shaoyuan and Kong, Lingdong and Dong, Yuhao and Sima, Chonghao and Zhang, Wenwei and Chen, Qi Alfred and Liu, Ziwei and Pan, Liang},
  booktitle={Proceedings of the IEEE/CVF International Conference on Computer Vision},
  pages={6585--6597},
  year={2025}
}

@inproceedings{marcu2024lingoqa,
  title={Lingoqa: Visual question answering for autonomous driving},
  author={Marcu, Ana-Maria and Chen, Long and H{\"u}nermann, Jan and Karnsund, Alice and Hanotte, Benoit and Chidananda, Prajwal and Nair, Saurabh and Badrinarayanan, Vijay and Kendall, Alex and Shotton, Jamie and others},
  booktitle={European Conference on Computer Vision},
  pages={252--269},
  year={2024},
  organization={Springer}
}

@inproceedings{lin2014microsoft,
  title={Microsoft coco: Common objects in context},
  author={Lin, Tsung-Yi and Maire, Michael and Belongie, Serge and Hays, James and Perona, Pietro and Ramanan, Deva and Doll{\'a}r, Piotr and Zitnick, C Lawrence},
  booktitle={European conference on computer vision},
  pages={740--755},
  year={2014},
  organization={Springer}
}

@article{dauner2024navsim,
  title={Navsim: Data-driven non-reactive autonomous vehicle simulation and benchmarking},
  author={Dauner, Daniel and Hallgarten, Marcel and Li, Tianyu and Weng, Xinshuo and Huang, Zhiyu and Yang, Zetong and Li, Hongyang and Gilitschenski, Igor and Ivanovic, Boris and Pavone, Marco and others},
  journal={Advances in Neural Information Processing Systems},
  volume={37},
  pages={28706--28719},
  year={2024}
}

@article{jia2024bench2drive,
  title={Bench2drive: Towards multi-ability benchmarking of closed-loop end-to-end autonomous driving},
  author={Jia, Xiaosong and Yang, Zhenjie and Li, Qifeng and Zhang, Zhiyuan and Yan, Junchi},
  journal={Advances in Neural Information Processing Systems},
  volume={37},
  pages={819--844},
  year={2024}
}

@article{sahoo2024simple,
  title={Simple and effective masked diffusion language models},
  author={Sahoo, Subham S and Arriola, Marianne and Schiff, Yair and Gokaslan, Aaron and Marroquin, Edgar and Chiu, Justin T and Rush, Alexander and Kuleshov, Volodymyr},
  journal={Advances in Neural Information Processing Systems},
  volume={37},
  pages={130136--130184},
  year={2024}
}

@article{bengio2015scheduled,
  title={Scheduled sampling for sequence prediction with recurrent neural networks},
  author={Bengio, Samy and Vinyals, Oriol and Jaitly, Navdeep and Shazeer, Noam},
  journal={Advances in neural information processing systems},
  volume={28},
  year={2015}
}

@inproceedings{ma2024dolphins,
  title={Dolphins: Multimodal language model for driving},
  author={Ma, Yingzi and Cao, Yulong and Sun, Jiachen and Pavone, Marco and Xiao, Chaowei},
  booktitle={European Conference on Computer Vision},
  pages={403--420},
  year={2024},
  organization={Springer}
}

@inproceedings{nie2024reason2drive,
  title={Reason2drive: Towards interpretable and chain-based reasoning for autonomous driving},
  author={Nie, Ming and Peng, Renyuan and Wang, Chunwei and Cai, Xinyue and Han, Jianhua and Xu, Hang and Zhang, Li},
  booktitle={European Conference on Computer Vision},
  pages={292--308},
  year={2024},
  organization={Springer}
}

@inproceedings{wang2025omnidrive,
  title={Omnidrive: A holistic vision-language dataset for autonomous driving with counterfactual reasoning},
  author={Wang, Shihao and Yu, Zhiding and Jiang, Xiaohui and Lan, Shiyi and Shi, Min and Chang, Nadine and Kautz, Jan and Li, Ying and Alvarez, Jose M},
  booktitle={Proceedings of the computer vision and pattern recognition conference},
  pages={22442--22452},
  year={2025}
}

@article{mao2023gpt,
  title={Gpt-driver: Learning to drive with gpt},
  author={Mao, Jiageng and Qian, Yuxi and Ye, Junjie and Zhao, Hang and Wang, Yue},
  journal={arXiv preprint arXiv:2310.01415},
  year={2023}
}

@inproceedings{shao2024lmdrive,
  title={Lmdrive: Closed-loop end-to-end driving with large language models},
  author={Shao, Hao and Hu, Yuxuan and Wang, Letian and Song, Guanglu and Waslander, Steven L and Liu, Yu and Li, Hongsheng},
  booktitle={Proceedings of the IEEE/CVF conference on computer vision and pattern recognition},
  pages={15120--15130},
  year={2024}
}

@article{jiang2024senna,
  title={Senna: Bridging large vision-language models and end-to-end autonomous driving},
  author={Jiang, Bo and Chen, Shaoyu and Liao, Bencheng and Zhang, Xingyu and Yin, Wei and Zhang, Qian and Huang, Chang and Liu, Wenyu and Wang, Xinggang},
  journal={arXiv preprint arXiv:2410.22313},
  year={2024}
}

@inproceedings{xu2025drivegpt4,
  title={Drivegpt4-v2: Harnessing large language model capabilities for enhanced closed-loop autonomous driving},
  author={Xu, Zhenhua and Bai, Yan and Zhang, Yujia and Li, Zhuoling and Xia, Fei and Wong, Kwan-Yee K and Wang, Jianqiang and Zhao, Hengshuang},
  booktitle={Proceedings of the Computer Vision and Pattern Recognition Conference},
  pages={17261--17270},
  year={2025}
}

@inproceedings{chen2025drivinggpt,
  title={Drivinggpt: Unifying driving world modeling and planning with multi-modal autoregressive transformers},
  author={Chen, Yuntao and Wang, Yuqi and Zhang, Zhaoxiang},
  booktitle={Proceedings of the IEEE/CVF International Conference on Computer Vision},
  pages={26890--26900},
  year={2025}
}

@article{li2025drivevlaw0,
  title={DriveVLA-W0: World models amplify data scaling law in autonomous driving},
  author={Li, Yingyan and Shang, Shuyao and Liu, Weisong and Zhan, Bing and Wang, Haochen and Wang, Yuqi and Chen, Yuntao and Wang, Xiaoman and An, Yasong and Tang, Chufeng and others},
  journal={arXiv preprint arXiv:2510.12796},
  year={2025}
}

@article{hoogeboom2021argmax,
  title={Argmax flows and multinomial diffusion: Learning categorical distributions},
  author={Hoogeboom, Emiel and Nielsen, Didrik and Jaini, Priyank and Forr{\'e}, Patrick and Welling, Max},
  journal={Advances in neural information processing systems},
  volume={34},
  pages={12454--12465},
  year={2021}
}

@article{li2022diffusion,
  title={Diffusion-lm improves controllable text generation},
  author={Li, Xiang and Thickstun, John and Gulrajani, Ishaan and Liang, Percy S and Hashimoto, Tatsunori B},
  journal={Advances in neural information processing systems},
  volume={35},
  pages={4328--4343},
  year={2022}
}

@article{gong2022diffuseq,
  title={Diffuseq: Sequence to sequence text generation with diffusion models},
  author={Gong, Shansan and Li, Mukai and Feng, Jiangtao and Wu, Zhiyong and Kong, LingPeng},
  journal={arXiv preprint arXiv:2210.08933},
  year={2022}
}

@article{shi2024simplified,
  title={Simplified and generalized masked diffusion for discrete data},
  author={Shi, Jiaxin and Han, Kehang and Wang, Zhe and Doucet, Arnaud and Titsias, Michalis},
  journal={Advances in neural information processing systems},
  volume={37},
  pages={103131--103167},
  year={2024}
}

@article{ou2024your,
  title={Your absorbing discrete diffusion secretly models the conditional distributions of clean data},
  author={Ou, Jingyang and Nie, Shen and Xue, Kaiwen and Zhu, Fengqi and Sun, Jiacheng and Li, Zhenguo and Li, Chongxuan},
  journal={arXiv preprint arXiv:2406.03736},
  year={2024}
}

@article{arriola2025block,
  title={Block diffusion: Interpolating between autoregressive and diffusion language models},
  author={Arriola, Marianne and Gokaslan, Aaron and Chiu, Justin T and Yang, Zhihan and Qi, Zhixuan and Han, Jiaqi and Sahoo, Subham Sekhar and Kuleshov, Volodymyr},
  journal={arXiv preprint arXiv:2503.09573},
  year={2025}
}

@article{von2025generalized,
  title={Generalized interpolating discrete diffusion},
  author={Von R{\"u}tte, Dimitri and Fluri, Janis and Ding, Yuhui and Orvieto, Antonio and Sch{\"o}lkopf, Bernhard and Hofmann, Thomas},
  journal={arXiv preprint arXiv:2503.04482},
  year={2025}
}

@article{zheng2025diffusion,
  title={Diffusion-based planning for autonomous driving with flexible guidance},
  author={Zheng, Yinan and Liang, Ruiming and Zheng, Kexin and Zheng, Jinliang and Mao, Liyuan and Li, Jianxiong and Gu, Weihao and Ai, Rui and Li, Shengbo Eben and Zhan, Xianyuan and others},
  journal={arXiv preprint arXiv:2501.15564},
  year={2025}
}

@article{zhang2026efficient,
  title={Efficient and Explainable End-to-End Autonomous Driving via Masked Vision-Language-Action Diffusion},
  author={Zhang, Jiaru and Gagvani, Manav and Cui, Can and Peng, Juntong and Zhang, Ruqi and Wang, Ziran},
  journal={arXiv preprint arXiv:2602.20577},
  year={2026}
}

@article{liang2025discrete,
  title={Discrete diffusion vla: Bringing discrete diffusion to action decoding in vision-language-action policies},
  author={Liang, Zhixuan and Li, Yizhuo and Yang, Tianshuo and Wu, Chengyue and Mao, Sitong and Nian, Tian and Pei, Liuao and Zhou, Shunbo and Yang, Xiaokang and Pang, Jiangmiao and others},
  journal={arXiv preprint arXiv:2508.20072},
  year={2025}
}

@article{wen2025llada,
  title={Llada-vla: Vision language diffusion action models},
  author={Wen, Yuqing and Li, Hebei and Gu, Kefan and Zhao, Yucheng and Wang, Tiancai and Sun, Xiaoyan},
  journal={arXiv preprint arXiv:2509.06932},
  year={2025}
}

@article{wen2025dvla,
  title={dvla: Diffusion vision-language-action model with multimodal chain-of-thought},
  author={Wen, Junjie and Zhu, Minjie and Liu, Jiaming and Liu, Zhiyuan and Yang, Yicun and Zhang, Linfeng and Zhang, Shanghang and Zhu, Yichen and Xu, Yi},
  journal={arXiv preprint arXiv:2509.25681},
  year={2025}
}

@article{liu2026mmada,
  title={MMaDA-VLA: Large Diffusion Vision-Language-Action Model with Unified Multi-Modal Instruction and Generation},
  author={Liu, Yang and Ding, Pengxiang and Jiang, Tengyue and Wang, Xudong and Song, Wenxuan and Lin, Minghui and Zhao, Han and Zhang, Hongyin and Zhuang, Zifeng and Zhao, Wei and others},
  journal={arXiv preprint arXiv:2603.25406},
  year={2026}
}

@article{chen2026dfm,
  title={DFM-VLA: Iterative Action Refinement for Robot Manipulation via Discrete Flow Matching},
  author={Chen, Jiayi and Song, Wenxuan and Chen, Shuai and Wang, Jingbo and Li, Zhijun and Li, Haoang},
  journal={arXiv preprint arXiv:2603.26320},
  year={2026}
}

@article{bai2025qwen3,
  title={Qwen3-vl technical report},
  author={Bai, Shuai and Cai, Yuxuan and Chen, Ruizhe and Chen, Keqin and Chen, Xionghui and Cheng, Zesen and Deng, Lianghao and Ding, Wei and Gao, Chang and Ge, Chunjiang and others},
  journal={arXiv preprint arXiv:2511.21631},
  year={2025}
}

@article{hinton2015distilling,
  title={Distilling the knowledge in a neural network},
  author={Hinton, Geoffrey and Vinyals, Oriol and Dean, Jeff},
  journal={arXiv preprint arXiv:1503.02531},
  year={2015}
}

@inproceedings{kim2016sequence,
  title={Sequence-level knowledge distillation},
  author={Kim, Yoon and Rush, Alexander M},
  booktitle={Proceedings of the 2016 conference on empirical methods in natural language processing},
  pages={1317--1327},
  year={2016}
}

@article{sanh2019distilbert,
  title={DistilBERT, a distilled version of BERT: smaller, faster, cheaper and lighter},
  author={Sanh, Victor and Debut, Lysandre and Chaumond, Julien and Wolf, Thomas},
  journal={arXiv preprint arXiv:1910.01108},
  year={2019}
}

@article{ranzato2015sequence,
  title={Sequence level training with recurrent neural networks},
  author={Ranzato, Marc'Aurelio and Chopra, Sumit and Auli, Michael and Zaremba, Wojciech},
  journal={arXiv preprint arXiv:1511.06732},
  year={2015}
}

@article{li2024llava,
  title={Llava-onevision: Easy visual task transfer},
  author={Li, Bo and Zhang, Yuanhan and Guo, Dong and Zhang, Renrui and Li, Feng and Zhang, Hao and Zhang, Kaichen and Zhang, Peiyuan and Li, Yanwei and Liu, Ziwei and others},
  journal={arXiv preprint arXiv:2408.03326},
  year={2024}
}

@article{wu2022trajectory,
  title={Trajectory-guided control prediction for end-to-end autonomous driving: A simple yet strong baseline},
  author={Wu, Penghao and Jia, Xiaosong and Chen, Li and Yan, Junchi and Li, Hongyang and Qiao, Yu},
  journal={Advances in Neural Information Processing Systems},
  volume={35},
  pages={6119--6132},
  year={2022}
}

@inproceedings{jia2023think,
  title={Think twice before driving: Towards scalable decoders for end-to-end autonomous driving},
  author={Jia, Xiaosong and Wu, Penghao and Chen, Li and Xie, Jiangwei and He, Conghui and Yan, Junchi and Li, Hongyang},
  booktitle={Proceedings of the IEEE/CVF Conference on Computer Vision and Pattern Recognition},
  pages={21983--21994},
  year={2023}
}

@inproceedings{jia2023driveadapter,
  title={Driveadapter: Breaking the coupling barrier of perception and planning in end-to-end autonomous driving},
  author={Jia, Xiaosong and Gao, Yulu and Chen, Li and Yan, Junchi and Liu, Patrick Langechuan and Li, Hongyang},
  booktitle={Proceedings of the IEEE/CVF International Conference on Computer Vision},
  pages={7953--7963},
  year={2023}
}

@article{zhai2023rethinking,
  title={Rethinking the open-loop evaluation of end-to-end autonomous driving in nuscenes},
  author={Zhai, Jiang-Tian and Feng, Ze and Du, Jinhao and Mao, Yongqiang and Liu, Jiang-Jiang and Tan, Zichang and Zhang, Yifu and Ye, Xiaoqing and Wang, Jingdong},
  journal={arXiv preprint arXiv:2305.10430},
  year={2023}
}

@inproceedings{yao2026drivesuprim,
  title={Drivesuprim: Towards precise trajectory selection for end-to-end planning},
  author={Yao, Wenhao and Li, Zhenxin and Lan, Shiyi and Wang, Zi and Sun, Xinglong and Alvarez, Jose M and Wu, Zuxuan},
  booktitle={Proceedings of the AAAI Conference on Artificial Intelligence},
  volume={40},
  number={14},
  pages={11910--11918},
  year={2026}
}

@inproceedings{hu2023planning,
  title={Planning-oriented autonomous driving},
  author={Hu, Yihan and Yang, Jiazhi and Chen, Li and Li, Keyu and Sima, Chonghao and Zhu, Xizhou and Chai, Siqi and Du, Senyao and Lin, Tianwei and Wang, Wenhai and others},
  booktitle={Proceedings of the IEEE/CVF conference on computer vision and pattern recognition},
  pages={17853--17862},
  year={2023}
}

@inproceedings{jiang2023vad,
  title={Vad: Vectorized scene representation for efficient autonomous driving},
  author={Jiang, Bo and Chen, Shaoyu and Xu, Qing and Liao, Bencheng and Chen, Jiajie and Zhou, Helong and Zhang, Qian and Liu, Wenyu and Huang, Chang and Wang, Xinggang},
  booktitle={Proceedings of the IEEE/CVF International Conference on Computer Vision},
  pages={8340--8350},
  year={2023}
}

@article{li2025recogdrive,
  title={Recogdrive: A reinforced cognitive framework for end-to-end autonomous driving},
  author={Li, Yongkang and Xiong, Kaixin and Guo, Xiangyu and Li, Fang and Yan, Sixu and Xu, Gangwei and Zhou, Lijun and Chen, Long and Sun, Haiyang and Wang, Bing and others},
  journal={arXiv preprint arXiv:2506.08052},
  year={2025}
}

@article{kim2026safedrive,
  title={SafeDrive: Fine-Grained Safety Reasoning for End-to-End Driving in a Sparse World},
  author={Kim, Jungho and Oh, Jiyong and Yu, Seunghoon and Shin, Hongjae and Kwak, Donghyuk and Choi, Jun Won},
  journal={arXiv preprint arXiv:2602.18887},
  year={2026}
}

@article{wang2025diffad,
  title={Diffad: A unified diffusion modeling approach for autonomous driving},
  author={Wang, Tao and Zhang, Cong and Qu, Xingguang and Li, Kun and Liu, Weiwei and Huang, Chang},
  journal={arXiv preprint arXiv:2503.12170},
  year={2025}
}

@article{wang2024qwen2,
  title={Qwen2-vl: Enhancing vision-language model's perception of the world at any resolution},
  author={Wang, Peng and Bai, Shuai and Tan, Sinan and Wang, Shijie and Fan, Zhihao and Bai, Jinze and Chen, Keqin and Liu, Xuejing and Wang, Jialin and Ge, Wenbin and others},
  journal={arXiv preprint arXiv:2409.12191},
  year={2024}
}

@article{tschannen2025siglip,
  title={Siglip 2: Multilingual vision-language encoders with improved semantic understanding, localization, and dense features},
  author={Tschannen, Michael and Gritsenko, Alexey and Wang, Xiao and Naeem, Muhammad Ferjad and Alabdulmohsin, Ibrahim and Parthasarathy, Nikhil and Evans, Talfan and Beyer, Lucas and Xia, Ye and Mustafa, Basil and others},
  journal={arXiv preprint arXiv:2502.14786},
  year={2025}
}

@misc{bai2025qwen25vltechnicalreport,
      title={Qwen2.5-VL Technical Report}, 
      author={Shuai Bai and Keqin Chen and Xuejing Liu and Jialin Wang and Wenbin Ge and Sibo Song and Kai Dang and Peng Wang and Shijie Wang and Jun Tang and Humen Zhong and Yuanzhi Zhu and Mingkun Yang and Zhaohai Li and Jianqiang Wan and Pengfei Wang and Wei Ding and Zheren Fu and Yiheng Xu and Jiabo Ye and Xi Zhang and Tianbao Xie and Zesen Cheng and Hang Zhang and Zhibo Yang and Haiyang Xu and Junyang Lin},
      year={2025},
      eprint={2502.13923},
      archivePrefix={arXiv},
      primaryClass={cs.CV},
      url={https://arxiv.org/abs/2502.13923}, 
}

@misc{zhu2025internvl3exploringadvancedtraining,
      title={InternVL3: Exploring Advanced Training and Test-Time Recipes for Open-Source Multimodal Models}, 
      author={Jinguo Zhu and Weiyun Wang and Zhe Chen and Zhaoyang Liu and Shenglong Ye and Lixin Gu and Hao Tian and Yuchen Duan and Weijie Su and Jie Shao and Zhangwei Gao and Erfei Cui and Xuehui Wang and Yue Cao and Yangzhou Liu and Xingguang Wei and Hongjie Zhang and Haomin Wang and Weiye Xu and Hao Li and Jiahao Wang and Nianchen Deng and Songze Li and Yinan He and Tan Jiang and Jiapeng Luo and Yi Wang and Conghui He and Botian Shi and Xingcheng Zhang and Wenqi Shao and Junjun He and Yingtong Xiong and Wenwen Qu and Peng Sun and Penglong Jiao and Han Lv and Lijun Wu and Kaipeng Zhang and Huipeng Deng and Jiaye Ge and Kai Chen and Limin Wang and Min Dou and Lewei Lu and Xizhou Zhu and Tong Lu and Dahua Lin and Yu Qiao and Jifeng Dai and Wenhai Wang},
      year={2025},
      eprint={2504.10479},
      archivePrefix={arXiv},
      primaryClass={cs.CV},
      url={https://arxiv.org/abs/2504.10479}, 
}

@misc{transfuser,
      title={TransFuser: Imitation with Transformer-Based Sensor Fusion for Autonomous Driving}, 
      author={Kashyap Chitta and Aditya Prakash and Bernhard Jaeger and Zehao Yu and Katrin Renz and Andreas Geiger},
      year={2022},
      eprint={2205.15997},
      archivePrefix={arXiv},
      primaryClass={cs.CV},
      url={https://arxiv.org/abs/2205.15997}, 
}

@article{feng2025artemis,
  title={Artemis: Autoregressive end-to-end trajectory planning with mixture of experts for autonomous driving},
  author={Feng, Renju and Xi, Ning and Chu, Duanfeng and Wang, Rukang and Deng, Zejian and Wang, Anzheng and Lu, Liping and Wang, Jinxiang and Huang, Yanjun},
  journal={IEEE Robotics and Automation Letters},
  volume={11},
  number={1},
  pages={226--233},
  year={2025},
  publisher={IEEE}
}

@article{englesson2021generalized,
  title={Generalized jensen-shannon divergence loss for learning with noisy labels},
  author={Englesson, Erik and Azizpour, Hossein},
  journal={Advances in Neural Information Processing Systems},
  volume={34},
  pages={30284--30297},
  year={2021}
}

@article{cen2024bridging,
  title={Bridging the training-inference gap in llms by leveraging self-generated tokens},
  author={Cen, Zhepeng and Liu, Yao and Zeng, Siliang and Chaudhari, Pratik and Rangwala, Huzefa and Karypis, George and Fakoor, Rasool},
  journal={arXiv preprint arXiv:2410.14655},
  year={2024}
}

@article{bachmann2024pitfalls,
  title={The pitfalls of next-token prediction},
  author={Bachmann, Gregor and Nagarajan, Vaishnavh},
  journal={arXiv preprint arXiv:2403.06963},
  year={2024}
}

@article{song2020lightpaff,
  title={LightPAFF: A two-stage distillation framework for pre-training and fine-tuning},
  author={Song, Kaitao and Sun, Hao and Tan, Xu and Qin, Tao and Lu, Jianfeng and Liu, Hongzhi and Liu, Tie-Yan},
  journal={arXiv preprint arXiv:2004.12817},
  year={2020}
}

@article{su2024roformer,
  title={Roformer: Enhanced transformer with rotary position embedding},
  author={Su, Jianlin and Ahmed, Murtadha and Lu, Yu and Pan, Shengfeng and Bo, Wen and Liu, Yunfeng},
  journal={Neurocomputing},
  volume={568},
  pages={127063},
  year={2024},
  publisher={Elsevier}
}

@article{ghosh2025pygraph,
  title={Pygraph: Robust compiler support for cuda graphs in pytorch},
  author={Ghosh, Abhishek and Nayak, Ajay and Panwar, Ashish and Basu, Arkaprava},
  journal={arXiv preprint arXiv:2503.19779},
  year={2025}
}

@article{ye2025flashinfer,
  title={Flashinfer: Efficient and customizable attention engine for llm inference serving},
  author={Ye, Zihao and Chen, Lequn and Lai, Ruihang and Lin, Wuwei and Zhang, Yineng and Wang, Stephanie and Chen, Tianqi and Kasikci, Baris and Grover, Vinod and Krishnamurthy, Arvind and others},
  journal={Proceedings of Machine Learning and Systems},
  volume={7},
  year={2025}
}

@article{hydramdp,
  title={Hydra-mdp: End-to-end multimodal planning with multi-target hydra-distillation},
  author={Li, Zhenxin and Li, Kailin and Wang, Shihao and Lan, Shiyi and Yu, Zhiding and Ji, Yishen and Li, Zhiqi and Zhu, Ziyue and Kautz, Jan and Wu, Zuxuan and others},
  journal={arXiv preprint arXiv:2406.06978},
  year={2024}
}

@misc{diffusiondrivev2,
      title={DiffusionDriveV2: Reinforcement Learning-Constrained Truncated Diffusion Modeling in End-to-End Autonomous Driving}, 
      author={Jialv Zou and Shaoyu Chen and Bencheng Liao and Zhiyu Zheng and Yuehao Song and Lefei Zhang and Qian Zhang and Wenyu Liu and Xinggang Wang},
      year={2025},
      eprint={2512.07745},
      archivePrefix={arXiv},
      primaryClass={cs.CV},
      url={https://arxiv.org/abs/2512.07745}, 
}

@misc{coco,
      title={Benchmarking Object Detectors with COCO: A New Path Forward}, 
      author={Shweta Singh and Aayan Yadav and Jitesh Jain and Humphrey Shi and Justin Johnson and Karan Desai},
      year={2024},
      eprint={2403.18819},
      archivePrefix={arXiv},
      primaryClass={cs.CV},
      url={https://arxiv.org/abs/2403.18819}, 
}

@inproceedings{yang2026drivemoe,
  title={Drivemoe: Mixture-of-experts for vision-language-action model in end-to-end autonomous driving},
  author={Yang, Zhenjie and Chai, Yilin and Jia, Xiaosong and Li, Qifeng and Shao, Yuqian and Zhu, Xuekai and Su, Haisheng and Yan, Junchi},
  booktitle={Proceedings of the IEEE/CVF Conference on Computer Vision and Pattern Recognition},
  pages={10678--10688},
  year={2026}
}

@article{li2026sgdrive,
  title={SGDrive: Scene-to-Goal Hierarchical World Cognition for Autonomous Driving},
  author={Li, Jingyu and Wu, Junjie and Hu, Dongnan and Huang, Xiangkai and Sun, Bin and Hao, Zhihui and Lang, Xianpeng and Zhu, Xiatian and Zhang, Li},
  journal={arXiv preprint arXiv:2601.05640},
  year={2026}
}

@inproceedings{zheng2026resad,
  title={Resad: Normalized residual trajectory modeling for end-to-end autonomous driving},
  author={Zheng, Zhiyu and Chen, Shaoyu and Yin, Haoran and Zhang, Xinbang and Zou, Jialv and Wang, Xinggang and Zhang, Qian and Zhang, Lefei},
  booktitle={Proceedings of the IEEE/CVF Conference on Computer Vision and Pattern Recognition},
  pages={3729--3739},
  year={2026}
}

@inproceedings{renz2025simlingo,
  title={Simlingo: Vision-only closed-loop autonomous driving with language-action alignment},
  author={Renz, Katrin and Chen, Long and Arani, Elahe and Sinavski, Oleg},
  booktitle={Proceedings of the Computer Vision and Pattern Recognition Conference},
  pages={11993--12003},
  year={2025}
}

\clearpage
\appendix
\section{Appendix}
\subsection{Training Datasets Construction}

To construct a unified, multi-task driving corpus, 
we aggregate and standardize multimodal data from five distinct sources. 
The resulting consolidated dataset comprises 981,361 samples for general multi-task learning, 
alongside dedicated splits for task-specific evaluations. 
Every sample is mapped to a standardized conversation format: 
an input containing visual streams and language context, 
and a target sequence representing actions, text, or structured coordinates.
\begin{itemize} 
\item NAVSIM~\cite{dauner2024navsim}: 
We utilize the official ``Navtest'' training split containing 103,288 samples. 
The inputs consist of front-view image, navigation targets, and ego-telemetry (velocity and acceleration), mapped to an 8-waypoint trajectory optimization target.

\item LingoQA~\cite{marcu2024lingoqa}:
To embed robust natural-language scene-reasoning and temporal video understanding, 
we incorporate 413,829 samples. 
Each instance has five sequentially ordered driving frames associated with an open-ended question regarding situational behaviors.

\item DriveLM~\cite{sima2024drivelm}:
We integrate 346,978 question-answering pairs across 4,072 frames to structurally enforce logical dependencies among perception, prediction, and planning tasks.

\item COCO~\cite{coco}: 
General spatial grounding and cross-modal object localization capabilities are preserved by incorporating 117,266 samples converted into a referring expression format.

\item SimLingo (Bench2Drive)~\cite{renz2025simlingo}: 
For closed-loop reactive planning, 
we utilize all 2,024,407 simulation frames from Bench2Drive. 
The configuration requires generating 6 future waypoints conditioned on structural camera tokens and ego-states.
\end{itemize}

\begin{table}[!h]
\centering
\caption{
Overview of the datasets used for training.
}
\label{tab:training_datasets}
\vspace{-2mm}
\renewcommand{\arraystretch}{1.12}
\small
\begin{tabular*}{\columnwidth}{
@{\extracolsep{\fill}}lccc@{}
}
\toprule
\textbf{Dataset}
& \textbf{Domain}
& \textbf{\# Samples}
& \textbf{Task} \\
\midrule
\multicolumn{4}{c}{\textit{Unified Multi-Task Split}} \\
\midrule
NAVSIM  & Driving & 103,288 & Planning \\
DriveLM & Driving & 346,978 & Driving VQA \\
LingoQA & Driving & 413,829 & Driving VQA \\
COCO-D  & Vision  & 117,266 & Detection \\
\midrule
\multicolumn{4}{c}{\textit{Task-Specific Splits}} \\
\midrule
SimLingo & Driving & 2,024,407 & Closed-loop planning \\
NAVSIM  & Driving & 103,288 & Open-loop planning \\
\bottomrule
\end{tabular*}
\end{table}

\subsection{Quantitative Evaluation Metrics}
To rigorously validate the multi-task and closed-loop capabilities of WAM-Diff2, 
we evaluate performance across five standard benchmarks spanning open/closed-loop motion planning, driving scene understanding, and spatial grounding.

\paragraph{Open-Loop Motion Planning.}
On the NAVSIM v1 benchmark, we report the official Predictive Driver Model Score (PDMS). 
The PDMS provides a holistic metric by evaluating a compound score of no at-fault collision ($\text{NC}$), 
drivable area compliance ($\text{DAC}$), 
time-to-collision ($\text{TTC}$), driving comfort ($\text{Comf.}$), 
and ego progress ($\text{EP}$):
\begin{equation}
\begin{aligned}
\mathrm{PDMS}
=
\mathrm{NC}\cdot\mathrm{DAC}\cdot
\frac{
5\,\mathrm{EP}
+
5\,\mathrm{TTC}
+
2\,\mathrm{Comf}
}{12}.
\end{aligned}
\label{eq:pdms}
\end{equation}

For the NAVSIM v2 benchmark, 
we adopt the Extended Predictive Driver Model Score (EPDMS). 
This metric introduces a broader set of constraints, 
including driving direction compliance ($\text{DDC}$), 
traffic light compliance ($\text{TLC}$), 
lane keeping ($\text{LK}$), history comfort ($\text{HC}$), 
and extended comfort ($\text{EC}$):
\begin{equation}
\begin{aligned}
\mathrm{EPDMS}
={}&
\mathrm{NC}\cdot\mathrm{DAC}\cdot
\mathrm{DDC}\cdot\mathrm{TLC} \\
&\cdot
\frac{
5\,\mathrm{EP}
+
5\,\mathrm{TTC}
+
2\,\mathrm{LK}
+
2\,\mathrm{HC}
+
2\,\mathrm{EC}
}{16}.
\end{aligned}
\label{eq:epdms}
\end{equation}

\paragraph{Closed-Loop Reactive Control.}
On the CARLA-driven Bench2Drive benchmark, 
we report the Success Rate (SR) and Driving Score (DS) to assess reactive planning under dynamic distribution shifts. 
Let $N_{\text{total}}$ be the total number of evaluated routes and $N_{\text{success}}$ be the number of successfully completed routes. 
The metrics are formulated as follows:
\begin{equation}
\mathrm{SR}
=\frac{N_{\mathrm{success}}}{N_{\mathrm{total}}},
\label{eq:success_rate}
\end{equation}
\begin{equation}
\mathrm{DS}
=\frac{1}{N_{\mathrm{total}}}
\sum_{i=1}^{N_{\mathrm{total}}}
\left(
\mathrm{RC}_{i}
\times
\prod_{j=1}^{M_i}
\mathrm{IS}_{i,j}
\right),
\label{eq:driving_score}
\end{equation}
where $\text{RC}_{i} \in [0, 100]$ represents the completion percentage of the $i$-th route, and $\text{IS}_{i,j} \in [0, 1)$ denotes the multiplicative penalty multiplier for the $j$-th infraction committed along that route.

\paragraph{Scene Reasoning and Spatial Grounding.}
For driving-oriented VQA, 
we evaluate open-ended situational scene reasoning via the Lingo-Judge metric on the LingoQA dataset and the GPT Score on the DriveBench dataset.
As for visual perception,
To quantify general cross-modal object localization and spatial grounding proficiency without task-specific heads, 
we evaluate structural JSON token predictions on the COCO dataset using standard mean Average Precision (mAP) at IoU thresholds from $0.50$ to $0.95$.

\subsection{Multi-Stage Optimization Protocol}

\paragraph{Model Architecture and Preprocessing.}
The framework is built upon the pre-trained Qwen3-VL-2B vision-language backbone. 
Camera inputs are uniformly scaled to $1920 \times 1080$ pixels and segmented into tokens using a spatial patch size of $16 \times 16$. 
The resulting visual embeddings are projected into the language model's hidden dimension space and concatenated with textual instruction tokens to establish a unified multimodal representation for downstream generation.

\paragraph{Training Phases and Objectives.}
The transformation from a sequential autoregressive generalist to a parallel discrete diffusion agent proceeds via a three-phase optimization pipeline:

\textit{Phase 1}: 
Multi-Task Autoregressive Pretraining. 
To establish a robust semantic foundation, the base model undergoes supervised fine-tuning for 5 epochs using a standard next-token prediction objective. 
All parameters are jointly optimized using the AdamW optimizer with a weight decay of 0.05, 
a cosine learning rate schedule, and a 10\% linear warmup. 
The base learning rate is fixed at $4 \times 10^{-5}$ under a global batch size of 128.

\textit{Phase 2}: 
Progressive Block-Causal Adaptation. 
To manage the severe architectural shift from sequential to parallel generation, we incrementally relax causal constraints by scaling the decoding block size:
\begin{equation}
\mathrm{AR} \rightarrow B_4 \rightarrow B_8 \rightarrow B_16 \rightarrow B_32.
\end{equation}
Each block variant is initialized with the weights of its predecessor ($\theta_{B}^{(0)} \leftarrow \theta_{B/2}$). 
For each configuration, the network is optimized for 5 epochs with a global batch size of 128. To stabilize cross-modal alignment during the attention shift, we apply a decoupled learning rate strategy: the language backbone is maintained at $4 \times 10^{-5}$, while the vision encoder is restricted to $2 \times 10^{-6}$.

\textit{Phase 3}: 
Hierarchical Knowledge Distillation.
To eliminate exposure bias and close the capacity gap resulting from compact parameterization, 
the 2B block-32 student model undergoes dual-stage distillation for 5 epochs. 
First, it is aligned with a stable 2B block-4 diffusion teacher (Stage II). 
Second, it inherits advanced cognitive reasoning from an 8B block-32 diffusion teacher (Stage III). 
Both stages optimize a symmetric Jensen-Shannon Divergence (JSD) loss over intermediate noisy states using a unified learning rate of $2 \times 10^{-6}$ across all components.

\paragraph{Hardware Infrastructure.}
The framework is implemented in PyTorch. Large-scale distributed training is executed across four compute nodes, 
each equipped with Kunpeng 920 7285Z processors and eight Ascend 910C NPUs., 
requiring a total allocation detailed in Table~\ref{tab:stage_training_hyperparameters}. 
Comprehensive inference throughput and closed-loop evaluations are performed on a single node utilizing eight Ascend 910C NPUs.
Hardware-specific inference optimizations are evaluated separately from the main experiments. In particular, the FlashInfer and CUDA Graph implementations are assessed only in the system-level efficiency study on a CUDA-compatible platform, whereas all model-quality results are obtained on the Ascend platform.

\begin{table}[!t]
\centering
\caption{Stage-wise training hyperparameters.}
\label{tab:stage_training_hyperparameters}
\vspace{-2mm}
\renewcommand{\arraystretch}{1.08}
\footnotesize

\begin{tabular*}{\columnwidth}{
    @{\extracolsep{\fill}}
    lccc
}
\toprule
\textbf{Hyperparameter}
& \textbf{Stage I}
& \textbf{Stage II}
& \textbf{Stage III} \\
\midrule

Epochs
& 5
& 5
& 5 \\

Batch size
& 128
& 128
& 128 \\

Learning rate
& $4{\times}10^{-5}$
& $2{\times}10^{-6}$
& $2{\times}10^{-6}$ \\

Weight decay
& 0.05
& 0.05
& 0.05 \\

Warmup ratio
& 0.1
& 0.1
& 0.1 \\

LR schedule
& Cosine
& Cosine
& Cosine \\

NPU hours
& 1,376
& 240
& 264 \\

\bottomrule
\end{tabular*}
\end{table}

\subsection{Multi-Task Prompt Templates}
\label{sec:prompt_templates}

We formulate all tasks using a unified vision--language instruction-following format. 
Each training sample is converted into a conversation consisting of a user message and an assistant response. 
The user message contains the visual input, task-specific context, and an explicit instruction, while the assistant response is used as the prediction target. 
We use the same prompt templates during training and evaluation, unless otherwise specified.

In the following templates, \texttt{<image>} denotes the visual placeholder processed by the vision encoder, and the contents enclosed by angle brackets denote sample-specific fields. 
Although the input prompts differ across tasks, all tasks share the same autoregressive or block-diffusion language modeling interface.

\paragraph{Motion Planning.}
For NAVSIM, the model receives front-view image together with navigation and ego-state information, including the current position, velocity, and acceleration. 
It is instructed to predict eight future waypoints covering the next four seconds.

\begin{quote}
\small\ttfamily
Here is front-view image from a driving vehicle: <image>

The navigation information is: <navigation>.

The current position is (<position\_x>, <position\_y>).

The current velocity is: (<velocity\_x>, <velocity\_y>) and the current acceleration is: (<acceleration\_x>, <acceleration\_y>).

Instruction: Based on the visual motion cues from the image (such as the relative speed of other vehicles and the changing distance to the intersection) and the provided telemetry, predict the optimal driving action for the next 4 seconds with 8 new waypoints.
\end{quote}

The target response is represented as a flattened sequence of two-dimensional coordinates:
\begin{quote}
\small\ttfamily
<x\_1>,<y\_1>,<x\_2>,<y\_2>,\ldots,<x\_8>,<y\_8>
\end{quote}

Similarly, for Bench2Drive, the model takes a front-view image, navigation targets (as a sequence of coordinates), a high-level command (e.g., "follow the road"), and the current scalar velocity. It is instructed to predict six future waypoints covering the next three seconds.

\begin{quote}
\small\ttfamily
Here is a front-view image from a driving vehicle:\\
\textless image\textgreater\\
The navigation information is: Target: [\textless target\_1\_x\textgreater, \textless target\_1\_y\textgreater] [\textless target\_2\_x\textgreater, \textless target\_2\_y\textgreater]. Command: \textless command\textgreater\\
Current velocity is: \textless current\_velocity\textgreater\\
Predict the optimal driving action for the next 3 seconds with 6 new waypoints.
\end{quote}

The target response follows the same flattened sequence structure, corresponding to the 6 predicted waypoints:
\begin{quote}
\small\ttfamily
\textless x\_1\textgreater,\textless y\_1\textgreater,\textless x\_2\textgreater,\textless y\_2\textgreater,\ldots,\textless x\_6\textgreater,\textless y\_6\textgreater
\end{quote}

No additional natural-language explanation is included in the planning target.

\paragraph{Driving Question Answering.}
We retain the original dataset-specific prompt formats for LingoQA and DriveBench.
For LingoQA, each sample consists of five temporally ordered driving frames followed by the original natural-language question:

\begin{quote}
\small\ttfamily
<image> <image> <image> <image> <image>\\
<question>
\end{quote}

For DriveBench, the input may contain either a single camera view or six surround-view images, with each image preceded by its camera identifier:

\begin{quote}
\small\ttfamily
[CAM\_VIEW] <image> \ldots\\
<question>
\end{quote}

DriveBench includes heterogeneous question types, such as action reasoning, collision-risk analysis, object-referenced reasoning, and multiple-choice ego-behavior prediction.
We preserve the original questions, camera identifiers, object references, and answer annotations for both datasets without rewriting.
The model directly generates the corresponding natural-language answer or selected option.

\paragraph{Visual Grounding.}
For COCO visual grounding, the model receives an image together with a referring expression using the following prompt:

\begin{quote}
\small\ttfamily
$<$image$>$\\
Locate every object that matches the description\\
``\{ref\_sentence\}'' in the image. Report bbox\\
coordinates in JSON format.
\end{quote}

The model is required to locate all objects matching the referring expression and return their bounding boxes and semantic labels in JSON format.
Each detected object is represented by a \texttt{bbox\_2d} field and a \texttt{label} field.
The bounding box is serialized in the coordinate order
$\left[y_{\min}, x_{\min}, y_{\max}, x_{\max}\right]$,
where all coordinates are normalized to a $1000 \times 1000$ coordinate space.
The target response follows the format:

\begin{quote}
\small\ttfamily
[\\
\hspace*{1em}\{\\
\hspace*{2em}"bbox\_2d": [$y_{\min}$, $x_{\min}$, $y_{\max}$, $x_{\max}$],\\
\hspace*{2em}"label": "$<$object\_class$>$"\\
\hspace*{1em}\},\\
\hspace*{1em}\ldots\\
]
\end{quote}

The output contains only the JSON-formatted detection results without any additional explanatory text.

\paragraph{Unified Output Interface.}
Despite the differences in task semantics, all targets are represented as token sequences and optimized with the same language-modeling objective. 
Motion planning and visual grounding use strictly formatted numerical outputs, whereas driving question-answering and scene-understanding tasks use natural-language outputs. 
During multi-task training, task identity is specified implicitly through the task-specific instruction and output format, without introducing additional task-specific prediction heads.

\subsection{Additional Qualitative Results}
\begin{figure}[!h]
    \centering
    \includegraphics[width=1.0\linewidth]{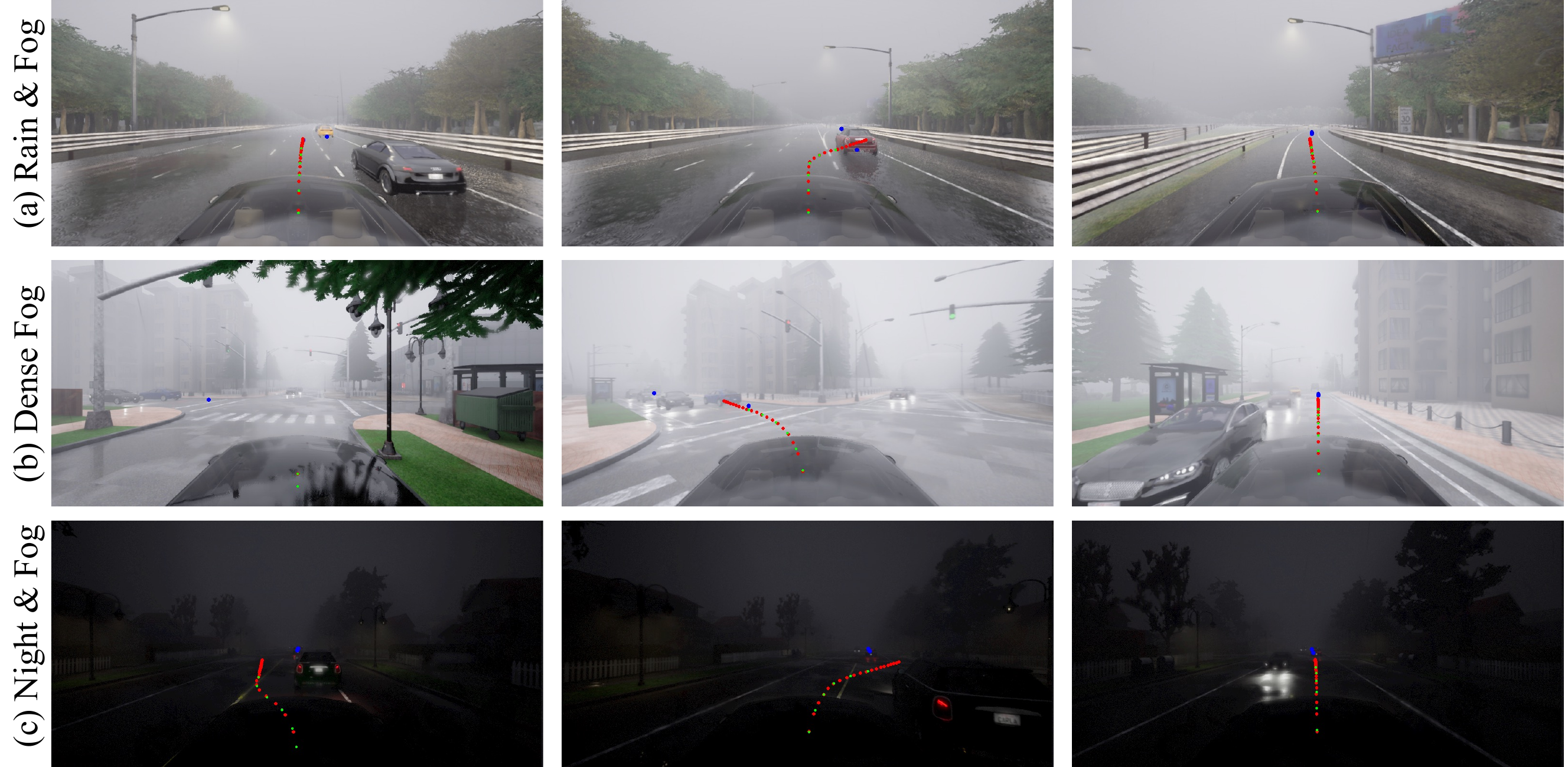}
    \vspace{-3mm}
    \caption{
        Representative extreme-weather driving scenarios on Bench2Drive, specifically featuring Night \& Fog, Dense Fog, and Rain \& Fog conditions.
    }
    \label{fig:extreme_weather}
\end{figure}

\begin{figure}[!h]
    \centering
    \includegraphics[width=1.0\linewidth]{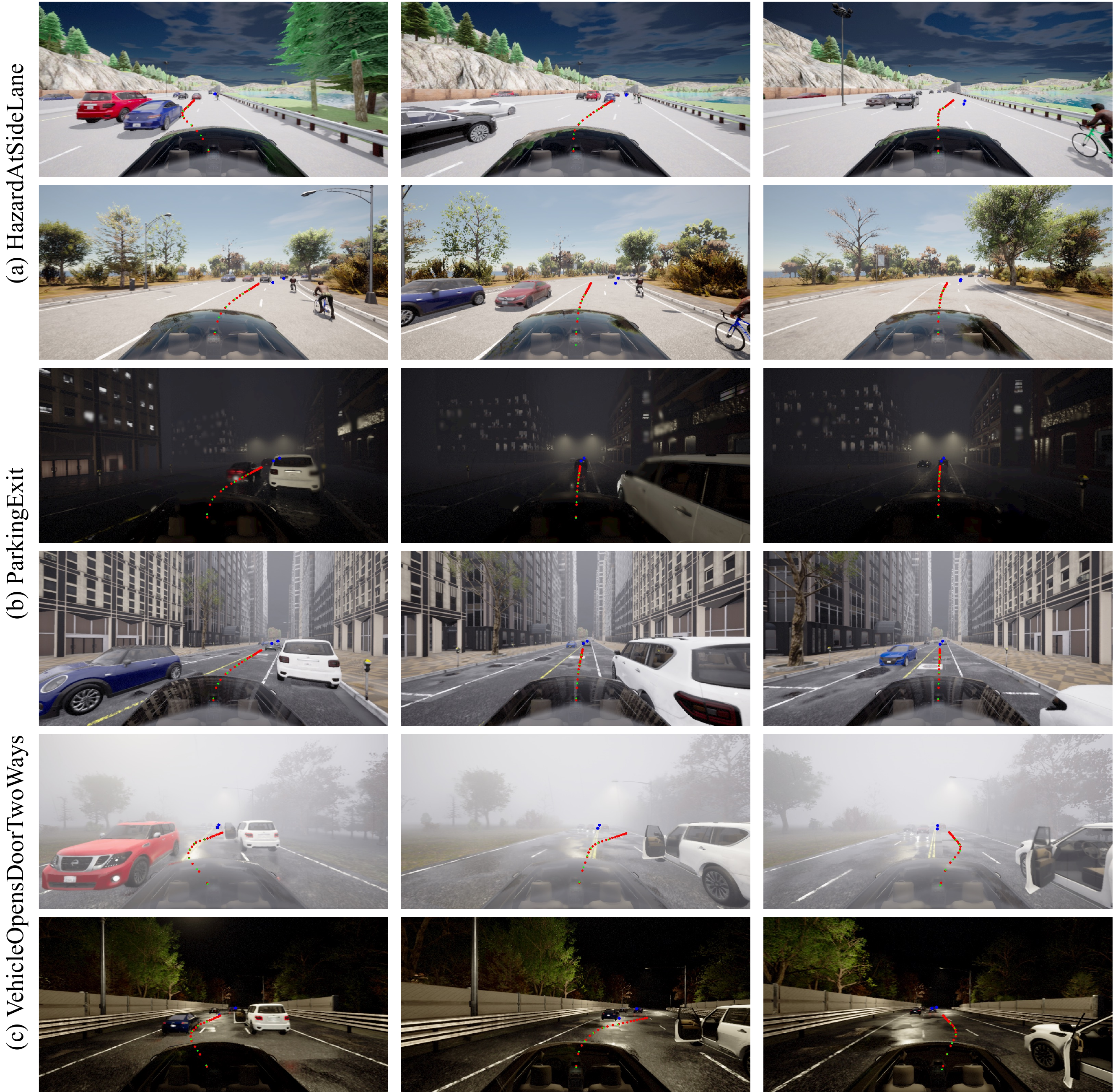}
    \vspace{-3mm}
    \caption{Representative driving scenarios in Bench2Drive under diverse weather and lighting conditions, including (a) HazardAtSideLane, (b) ParkingExit, and (c) VehicleOpensDoorTwoWays}.
    \label{fig:scenarios_of_b2d}
\end{figure}

We present extensive qualitative visualizations of WAMDiff2 on NAVSIM and Bench2Drive to demonstrate the effectiveness of our proposed
method (see Figure~\ref{fig:extreme_weather} and~\ref{fig:scenarios_of_b2d}).

\end{document}